\PassOptionsToPackage{table}{xcolor}
\documentclass[color=00935A]{seedstyle}
\usepackage{wrapfig}
\usepackage{dashrule}
\usepackage{mathtools}
\usepackage{pifont}
\usepackage{tabularx}
\usepackage{makecell}
\usepackage{flafter}
\usepackage{placeins}
\graphicspath{{images/}}
\newcommand{\benchmark}{\textsc{EngiWorld}}

\definecolor{taskblue}{HTML}{EAF1FA}
\definecolor{taskteal}{HTML}{E8F3EF}
\definecolor{taskpurple}{HTML}{F0ECF7}
\definecolor{taskgold}{HTML}{FAF1DF}
\definecolor{taskrose}{HTML}{F8EBEE}

\newcommand{\modellogo}[1]{\raisebox{-0.15ex}{\includegraphics[width=1em,height=1em,keepaspectratio]{images/model_logos/#1.pdf}}\hspace{0.35em}}
\newcommand{\appendixtablestyle}{\small\setlength{\tabcolsep}{3.5pt}\renewcommand{\arraystretch}{1.05}}
\usepackage{needspace} % keeps appendix section headings with their first lines

\teamname{Zhiman AI}
\teamlogo{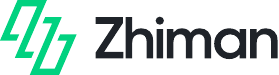}             % top-left logo
\partnerlogo{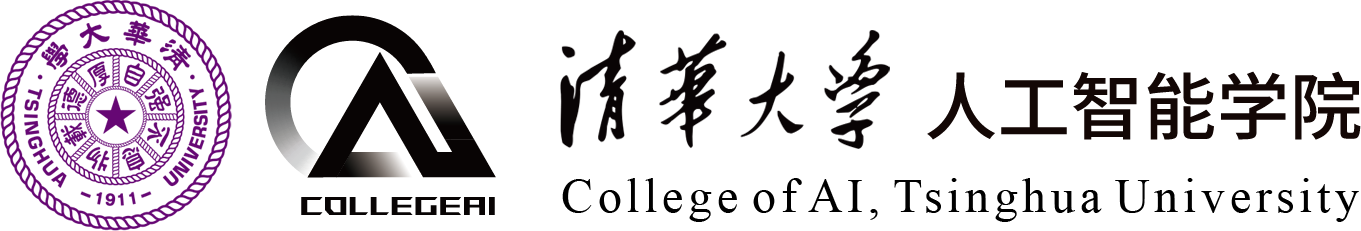}   % top-right logo

\title{EngiWorld: What Can Frontier Agents Deliver in Professional Engineering Environments?}

\author[1,2,*,\dagger]{Hongcheng Gao}
\author[3,*]{Hailong Qu}
\author[4,*]{Yu Lei}
\author[5]{Henghui Sun}
\author[6]{Haoyang Li}
\author[4]{Yipeng Wei}
\author[7]{Naihao Xue}
\author[8]{Xiaohan Yu}
\author[4]{Zhuo Tao}
\author[9]{Yihe Zang}
\author[4]{Yajiao Wang}
\author[10]{Jingyi Tang}
\author[4]{Yi Li}
\author[4]{Jingjing Zhou}
\author[2]{Jie Luo}
\author[10]{Bohan Zeng}
\author[10]{Chengyu Shen}
\author[11]{Hao Jiang}
\author[1]{Chong Chen}
\author[10]{Bowen Qu}
\author[10]{Olive Huang}
\author[2]{Zeqiang Wang}

\affiliation[1]{Tsinghua University\quad $^{2}$Zhiman AI\quad $^{3}$Chongqing University\quad $^{4}$UCAS\quad $^{5}$Shandong University}
\affiliation[6]{BUPT\quad $^{7}$Fudan University\quad $^{8}$Henan Polytechnic University\quad $^{9}$Xi'an Jiaotong University}
\affiliation[10]{Peking University\quad $^{11}$Zhejiang University}

\contribution[*]{Equal contribution}
\contribution[\dagger]{Corresponding author}

\abstract{% !TEX root = ../main.tex
Autonomous agents have made rapid progress in general-purpose computer use, yet reliable automation of professional industrial engineering remains out of reach. Engineering workflows require reasoning over geometric and physical constraints and keeping dependencies consistent across software and design stages. We present \textbf{\benchmark}, a benchmark designed around the capabilities that real engineering projects require, from operating a single application to solving engineering problems end to end. It contains 1,301 expert-curated tasks spanning six engineering domains (CAD, CAE, CAM, BIM, EDA, and DCC) and 26 professional software applications, with both GUI and CLI interfaces. The tasks are organized into six types, corresponding to capabilities that real engineering projects require. We further introduce an artifact-centric evaluation methodology built on a unified suite of domain verifiers. The verifiers programmatically check whether final and intermediate artifacts are geometrically valid, physically feasible, and compliant with design rules. Beyond binary success, they also score how well optimized designs meet their specifications. Evaluation of seven frontier models reveals a substantial capability gap: the strongest model scores only 44.7/100, and just 3.6\% of runs on tasks that require cross-software coordination succeed. \benchmark{} provides a rigorous foundation for measuring progress toward agents that operate professional engineering software end to end.
}

\date{\today}
\checkdata[Project Page]{\url{https://engiworld.github.io}}

\begin{document}
\maketitle

% !TEX root = ../main.tex
% \vspace{-0.1cm}
\section{Introduction}
% \vspace{-0.1cm}
\label{sec:introduction}

Large language models (LLMs), multimodal large language models (MLLMs), and agents built on them have made transformative progress on general computer-use tasks, and can now plan, perceive, and act across digital environments spanning web navigation, office productivity, and software engineering~\citep{deng2023mind2web,zhou2023webarena,xie2024osworld,rawles2024androidworld,kapoor2024omniact,jimenez2023swe,yang2023intercode,huang2023mlagentbench,lai2023ds}. However, their potential in professional industrial design software, a domain of enormous economic and societal value, remains largely unexplored~\citep{ren2025industrial,gao2025lifecycle,liu2026idesigngpt}.

\begin{figure}[t]
% \vspace{-0.2cm}
    \centering
    \includegraphics[width=\linewidth]{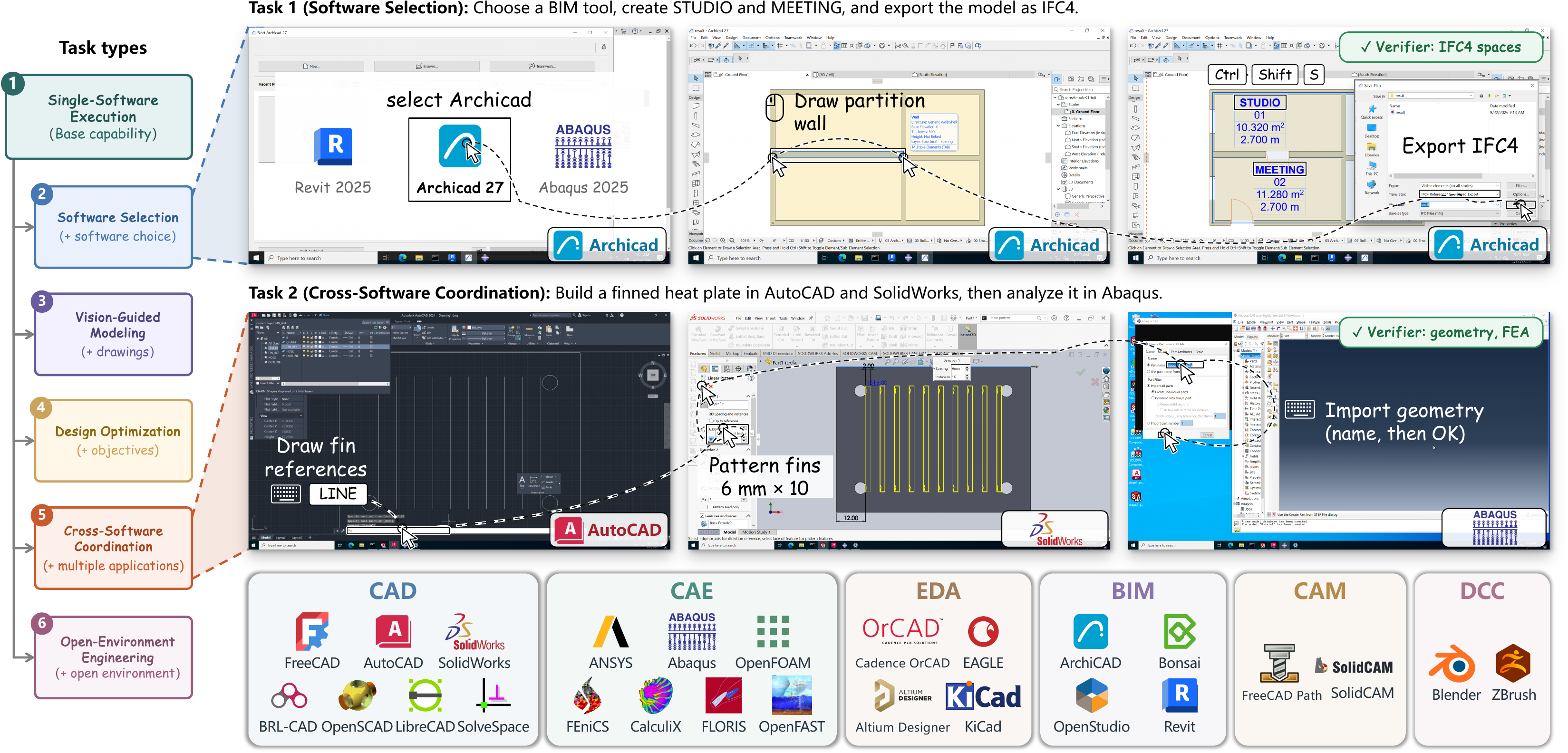}
        % \vspace{-0.2cm}
    \caption{\textbf{Overview of \benchmark{}.} Six task types across six engineering domains evaluate the capabilities required throughout a real engineering workflow, from operating one application to solving open-ended problems.}
    \label{fig:engiworld-overview}
    % \vspace{-0.2cm}
\end{figure}

Although existing benchmarks evaluate general digital workflows well, they show three major limitations in industrial domains.
\textbf{First}, lack of artifact verification. Benchmarks such as OSWorld~\citep{xie2024osworld}, Spider2-V~\citep{cao2024spider2v}, and ScienceBoard~\citep{sun2026scienceboard} check output files and application states against reference files or expected values. Such checks suit office and data workflows but cannot establish whether an engineering artifact is valid: an engineering model that looks correct may still be unusable because of non-manifold geometry or improper topological constraints~\citep{wu2021deepcad,jayaraman2021uv,li2022free2cad}. FEABench~\citep{mudur2025feabench}, BIM-Edit~\citep{nithyanantham2026bimedit}, and CADTestBench~\mbox{\citep{mallis2026cadtests}} have begun to validate artifact quality, but cover only a single software application and a single task format.
\textbf{Second}, lack of long-horizon interaction. Text2CAD~\citep{khan2024text2cad} focuses on single-turn generation, and although CADWorld~\citep{dong2026cadworld} introduces GUI interaction, its workflows stay within a single application (FreeCAD). But industrial tasks require maintaining parametric logic over long operation sequences and across applications, where an early error can invalidate all subsequent operations~\citep{man2025videocad,gong2026toolcad}.
\textbf{Third}, lack of native software feedback. Tool-using engineering agents such as ChatEDA~\citep{wu2024chateda} and Foam-Agent~\citep{yue2025foam2} observe only textual signals such as execution logs and error messages, and GUI-EDA~\citep{li2025guieda} even resorts to offline single-step prediction without executing any action. In contrast, engineers rely on native feedback from domain software, such as rendered outputs, simulation convergence curves, and design-rule violation reports, to iteratively refine their designs.

More fundamentally, a real engineering project requires a range of capabilities from an engineer. The basic capability is to operate professional software to build, modify, and simulate designs. On this basis, the engineer must also choose suitable software for a requirement, read design intent from engineering drawings, improve a design against quantitative objectives under engineering constraints, coordinate work across several applications when one application cannot meet the requirement, and solve open-ended problems on a computer without preinstalled engineering software. Yet existing benchmarks cover only part of these capabilities.

To bridge this gap, we introduce \textbf{\benchmark}, a benchmark designed around these capabilities for evaluating agents in professional industrial design environments. As shown in Figure~\ref{fig:engiworld-overview}, \benchmark{} spans six engineering domains, namely computer-aided design (CAD), computer-aided engineering (CAE), computer-aided manufacturing (CAM), building information modeling (BIM), electronic design automation (EDA), and digital content creation (DCC), and contains 1,301 expert-curated tasks across 26 software applications. The tasks are grounded in real engineering processes~\citep{gao2025lifecycle} and require agents to interact with real professional software over multiple turns through GUI or CLI interfaces. The tasks in \benchmark{} are organized into six types. \emph{Single-Software Execution} tests the basic capability of operating one application. The other five types go beyond it: \emph{Software Selection} tests choosing suitable software, \emph{Vision-Guided Modeling} tests working from engineering drawings, \emph{Design Optimization} tests optimizing a design under constraints, \emph{Cross-Software Coordination} tests coordinating several applications, and \emph{Open-Environment Engineering} tests solving problems on a computer without preinstalled engineering software (Section~\ref{sec:task-formulation}).

Furthermore, we propose an \textbf{artifact-centric evaluation methodology}. Because industrial artifacts rarely have a unique correct answer, we do not compare them with a single reference. Instead, we extend artifact-level programmatic verification from single-domain benchmarks~\citep{mudur2025feabench,nithyanantham2026bimedit,mallis2026cadtests,singh2026cadengbench} into a unified suite of domain verifiers spanning all six domains, which check the geometric accuracy, physical feasibility, and rule compliance of both final and intermediate artifacts (e.g., STEP files, netlists, and G-code). Unlike most interactive benchmarks, which judge only success or failure, \benchmark{} assigns continuous scores to Design Optimization tasks: designs first pass feasibility checks and then receive quality scores according to how well they meet their specifications. We therefore evaluate not only \emph{whether} a task is completed but also \emph{how well} it is completed.

We evaluate seven frontier models, five proprietary and two open-source. Even the strongest model achieves a \textbf{score of only 44.7/100}, and all models perform worse on GUI tasks than on CLI tasks. Coordination across software is the main bottleneck: agents rarely complete Cross-Software Coordination tasks. Most failed runs either declare completion with artifacts that fail verification or reach the turn limit. These findings highlight the profound gap between general intelligence and the demands of professional engineering, positioning \benchmark{} as a rigorous foundation for future research on industrial agents.
% !TEX root = ../main.tex
\section{Related Work}
\label{sec:related-work}

\begingroup
\providecommand{\cmark}{\textcolor{green!60!black}{\ding{51}}}
\providecommand{\xmark}{\textcolor{red!75!black}{\ding{55}}}
\providecommand{\pmark}{\textcolor{orange!85!black}{\ensuremath{\circ}}}
\begin{table}[!h]
\centering
\caption{\textbf{Comparison with existing benchmarks.} \textit{Software}: number of software applications. \textit{Interface}: how agents operate the software; GUI (offline) denotes action prediction without execution. \textit{Cross}: multi-software workflows. \textit{Select.}: agents choose the software. \textit{Optim.}: continuous scoring of design objectives. \textit{Verif.}: programmatic checks of engineering properties. \pmark: partial support. ALE (Eng.): the subset of 25 engineering and DCC tasks in ALE.}
\label{tab:industrial-benchmarks}
\begingroup
\setlength{\tabcolsep}{2pt}
\renewcommand{\arraystretch}{1.0}
\footnotesize
\begin{tabular}{@{}l l@{\hspace{1pt}}c@{\hspace{3pt}}c c@{\hspace{3pt}}c@{\hspace{3pt}}c@{\hspace{3pt}}c@{\hspace{3pt}}c@{}}
\toprule
\textbf{Benchmark} & \textbf{Scale} & \textbf{Software} & \textbf{Domain} & \textbf{Interface} & \textbf{Cross} & \textbf{Select.} & \textbf{Optim.} & \textbf{Verif.} \\
\midrule
\multicolumn{9}{@{}l}{\textit{Engineering artifact generation}} \\
Text2CAD~\citep{khan2024text2cad} & 170k models & -- & CAD & -- & \xmark & \xmark & \xmark & \xmark \\
VerilogEval~\citep{liu2023verilogeval} & 156 problems & -- & EDA & -- & \xmark & \xmark & \xmark & \cmark \\
RTLLM~\citep{lu2024rtllm} & 30 designs & -- & EDA & -- & \xmark & \xmark & \cmark & \cmark \\
CADTestBench~\citep{mallis2026cadtests} & 200 designs & -- & CAD & -- & \xmark & \xmark & \xmark & \cmark \\
CADEngBench~\citep{singh2026cadengbench} & 600 tasks & -- & CAD/CAE & -- & \xmark & \xmark & \xmark & \cmark \\
\midrule
\multicolumn{9}{@{}l}{\textit{General computer-use agents}} \\
OSWorld~\citep{xie2024osworld} & 369 tasks & 9 & Desktop & GUI+CLI & \cmark & \xmark & \xmark & \xmark \\
OSWorld 2.0~\citep{yuan2026osworld2} & 108 tasks & 18 & Desktop & GUI+CLI & \cmark & \xmark & \xmark & \xmark \\
Spider 2.0~\citep{lei2024spider2} & 632 tasks & 7 & Data eng. & CLI & \xmark & \xmark & \xmark & \xmark \\
Spider2-V~\citep{cao2024spider2v} & 494 tasks & 20 & Data eng. & GUI+CLI & \cmark & \xmark & \xmark & \xmark \\
ScienceBoard~\citep{sun2026scienceboard} & 169 tasks & 6 & Science & GUI+CLI & \cmark & \xmark & \xmark & \xmark \\
ALE (Eng.)~\citep{sun2026agentsexam} & 25 tasks & 13 & Eng./DCC & GUI+CLI & \cmark & \xmark & \xmark & \pmark \\
\midrule
\multicolumn{9}{@{}l}{\textit{Engineering software agents}} \\
VideoCAD~\citep{man2025videocad} & 41k videos & 1 & CAD & GUI (offline) & \xmark & \xmark & \xmark & \xmark \\
GUI-EDA~\citep{li2025guieda} & 2,082 steps & 5 & EDA/CAE & GUI (offline) & \xmark & \xmark & \xmark & \xmark \\
FEABench~\citep{mudur2025feabench} & 215 tasks & 1 & CAE & CLI & \xmark & \xmark & \xmark & \cmark \\
BlenderGym~\citep{gu2025blendergym} & 245 pairs & 1 & DCC & CLI & \xmark & \xmark & \xmark & \xmark \\
CADWorld~\citep{dong2026cadworld} & 200 tasks & 1 & CAD/CAE/CAM & GUI & \xmark & \xmark & \xmark & \cmark \\
BIM-Edit~\citep{nithyanantham2026bimedit} & 324 tasks & 1 & BIM & CLI & \xmark & \xmark & \xmark & \cmark \\
\midrule
\textbf{\benchmark{}} & \textbf{1,301 tasks} & \textbf{26} & \makecell{\textbf{CAD/CAE/CAM/}\\\textbf{BIM/EDA/DCC}} & \textbf{GUI+CLI} & \cmark & \cmark & \cmark & \cmark \\
\bottomrule
\end{tabular}%
\endgroup
\end{table}
\endgroup

\noindent\textbf{Computer-Use Benchmarks.}
Benchmarks for autonomous agents cover software engineering~\citep{jimenez2023swe,yang2023intercode,huang2023mlagentbench}, web interaction~\citep{deng2023mind2web,zhou2023webarena,koh2024visualwebarena}, and desktop operation~\citep{xie2024osworld}. OSWorld evaluates task completion in real desktop environments. Spider 2.0~\citep{lei2024spider2} evaluates enterprise data workflows through SQL and code, and Spider2-V~\citep{cao2024spider2v} and ScienceBoard~\citep{sun2026scienceboard} extend evaluation to professional data-engineering and scientific workflows. These settings combine interaction through graphical or command-line interfaces with execution-based outcome checks. More recent benchmarks examine long-horizon execution and workflow dependencies: OSWorld 2.0~\citep{yuan2026osworld2} extends OSWorld to long-horizon tasks across multiple applications, and Agents' Last Exam (ALE)~\citep{sun2026agentsexam} evaluates professional workflows with verifiable deliverables across industries, including engineering and 3D applications. These benchmarks provide realistic interactive environments, but their checks rarely examine the engineering properties of the produced artifacts.

\noindent\textbf{Engineering Agents and Benchmarks.}
Engineering automation encompasses both structured artifact generation and interactive software use. DeepCAD~\citep{wu2021deepcad} and Text2CAD~\citep{khan2024text2cad} generate parametric CAD construction sequences, while RTLCoder~\citep{liu2024rtlcoder} generates hardware descriptions. VerilogEval~\citep{liu2023verilogeval} assesses functional correctness through simulation, and RTLLM~\citep{lu2024rtllm} additionally evaluates power, performance, and area objectives. Tool-using approaches incorporate planning and execution feedback into CAD modeling, electronic design, and numerical simulation~\citep{gong2026toolcad,wu2024chateda,chen2024metaopenfoam,yue2025foam2}. 
% VideoCAD~\citep{man2025videocad} and GUI-EDA~\citep{li2025guieda} provide datasets for learning professional GUI interactions and evaluating action prediction and visual understanding.
% VideoCAD~\citep{man2025videocad} and GUI-EDA~\citep{li2025guieda} provide datasets for professional GUI interaction learning and evaluation for action prediction.
VideoCAD~\citep{man2025videocad} and GUI-EDA~\citep{li2025guieda} provide datasets of professional GUI interactions for training and evaluation.

Engineering benchmarks also assess whether generated artifacts satisfy domain-specific requirements. CADWorld~\citep{dong2026cadworld} evaluates GUI-based FreeCAD workflows spanning modeling, simulation, and manufacturing. CADTestBench~\citep{mallis2026cadtests} uses executable tests to check geometric and topological requirements, while CADEngBench~\citep{singh2026cadengbench} evaluates parametric part generation, assembly reasoning, and linear-static finite element analysis with validity and design-for-manufacturing checks. Beyond CAD, FEABench~\citep{mudur2025feabench} evaluates multiphysics problem solving in COMSOL, BIM-Edit~\citep{nithyanantham2026bimedit} assesses edits to Industry Foundation Classes (IFC) models through geometric, semantic, and topological criteria, and BlenderGym~\citep{gu2025blendergym} evaluates graphics editing through start--goal scene pairs. Across these benchmarks, evaluation ranges from functional simulation and executable constraint checks to geometric and visual comparisons, reflecting the different requirements of engineering artifacts. As Table~\ref{tab:industrial-benchmarks} shows, none of these engineering benchmarks supports cross-software workflows or lets agents select software, and only RTLLM assigns continuous scores to design objectives. In contrast, \benchmark{} combines these capabilities with programmatic verification across all six engineering domains.

% !TEX root = ../main.tex
% \vspace{-0.2cm}
\section{The \texorpdfstring{\benchmark{}}{EngiWorld} Benchmark}
% \vspace{-0.2cm}
\label{sec:benchmark}

In this section, we provide the benchmark overview and task formulation (Section~\ref{sec:task-formulation}), the software environments and interaction interfaces (Section~\ref{sec:interaction}), the task construction pipeline and benchmark coverage (Section~\ref{sec:construction}), and the artifact-centric evaluation protocol (Section~\ref{sec:verification}).

\subsection{Benchmark Overview and Task Formulation}
\label{sec:task-formulation}

An \benchmark{} task places an agent in a computer environment and asks it to produce or modify engineering artifacts. Tasks are organized into six types. The first type tests the basic capability of operating one application. The other five types go beyond it and test other capabilities that real engineering projects require: \textbf{(1) Single-Software Execution}: model, modify, or simulate a design in designated software; \textbf{(2) Software Selection}: choose suitable software from a set of candidates and complete the task with it; \textbf{(3) Vision-Guided Modeling}: build an engineering model from reference drawings, which requires spatial and geometric reasoning to fully understand the actual engineering requirements; \textbf{(4) Design Optimization}: improve a design metric, such as part mass or PCB layout cost, under engineering constraints; \textbf{(5) Cross-Software Coordination}: carry out a long-horizon workflow that one application cannot complete, passing artifacts across several applications so that each stage builds on the previous one; \textbf{(6) Open-Environment Engineering}: solve a problem end to end on a computer without preinstalled engineering software, choosing and installing the software and planning the solution with no constraints on software, order, or intermediate artifacts. Only Vision-Guided Modeling tasks include reference drawings; the other task types give requirements in text.

\begin{figure}[!htbp]
    \centering
    \includegraphics[width=0.95\linewidth]{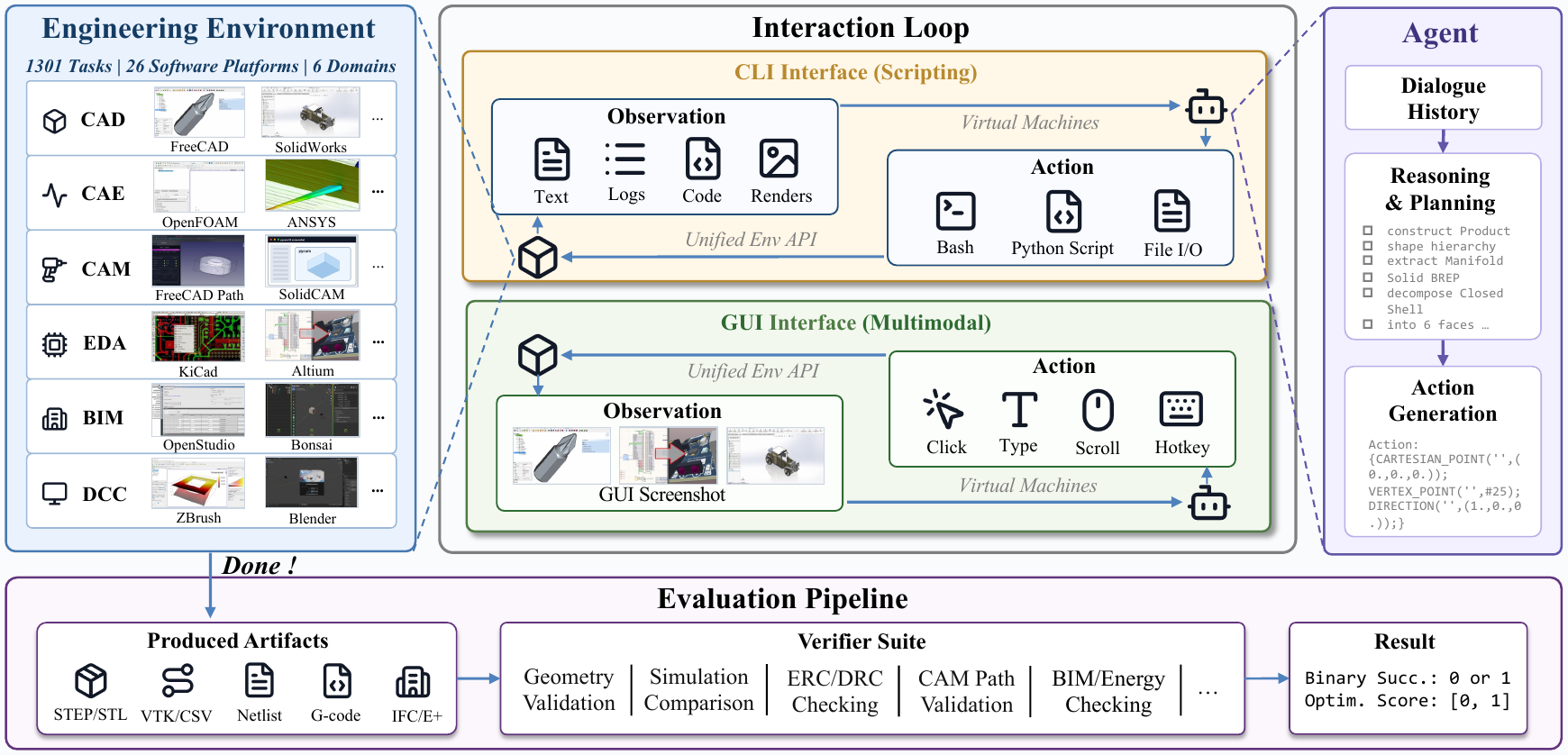}
    \caption{\textbf{Overview of the \benchmark{} framework.} Agents operate engineering software through CLI or GUI interfaces, and domain-specific verifiers check whether the resulting artifacts satisfy the engineering requirements.}
    \label{fig:engiworld-framework}
\end{figure}

Each task $\tau_i$ specifies an objective $g_i$, an initial workspace with input files (and reference drawings for Vision-Guided Modeling tasks), an action space $\mathcal{A}_i$ defined by the interface, and acceptance criteria. We model the interaction as a partially observable sequential decision process~\citep{kaelbling1998planning}. At turn $t$, the agent receives an observation $o_t$ (a screenshot for GUI tasks or terminal output for CLI tasks) and selects
\[
    a_t \sim \pi_\theta(\cdot\mid g_i,o_t,h_t),
    \qquad
    a_t\in\mathcal{A}_i\cup\{\texttt{DONE},\texttt{FAIL}\},
\]
where $h_t$ is the interaction history. Each action changes the software state and therefore the feedback available for the next decision. The episode ends at turn $T$ when the agent declares \texttt{DONE} or \texttt{FAIL} or reaches the turn limit. During the episode, the agent produces artifacts such as CAD models and simulation results, and these artifacts form the submission $\mathcal{Y}_T$. Figure~\ref{fig:engiworld-framework} summarizes this interaction and evaluation process.

\subsection{Environments and Interaction}
\label{sec:interaction}

\benchmark{} tasks run in isolated Windows or Ubuntu environments containing the required native engineering software, except for Open-Environment Engineering tasks. Each episode starts from a task-specific workspace state with the necessary input files prepared and the relevant documents opened. Single-Software Execution tasks use designated software, whereas Cross-Software Coordination tasks provide a shared workspace in which intermediate artifacts are transferred across software and used by subsequent stages. Software Selection tasks let the agent choose suitable software from several candidates.

The benchmark supports two interaction interfaces. With the GUI interface, agents operate the software through visual observations and mouse or keyboard actions. With the CLI interface, agents use terminal commands and scripts and receive textual feedback from the software and operating system. The two interfaces expose different observation and action spaces but share the same artifact-based evaluation protocol. Open-Environment Engineering tasks use the CLI interface because it is more flexible: agents can install software and drive it through commands, scripts, or GUI automation.

% \vspace{-0.1cm}
\subsection{Task Construction and Coverage}
% \vspace{-0.2cm}
\label{sec:construction}

\begin{figure}[t]
    \centering
    \includegraphics[width=\linewidth]{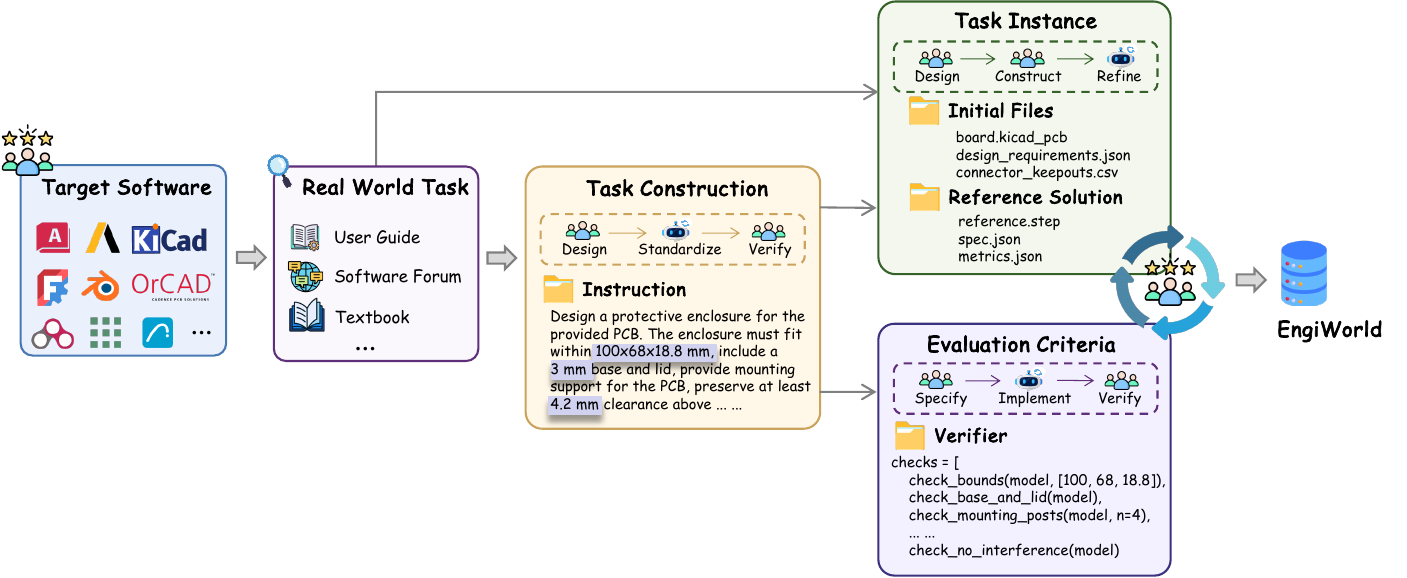}
    \caption{\textbf{Dataset construction pipeline.} Experts design and verify the instruction, task instance, and evaluation criteria, while AI standardizes and implements them. Each task enters \benchmark{} only after a final expert review.}
    \label{fig:dataset-construction}
\end{figure}

\begin{figure}[!htbp]
% \vspace{-0.2cm}
    \centering
    \captionsetup[subfigure]{font=footnotesize,labelfont=bf,justification=centering,singlelinecheck=false,skip=1pt}
    \begin{subfigure}[t]{0.595\linewidth}
        \centering
        \includegraphics[width=\linewidth]{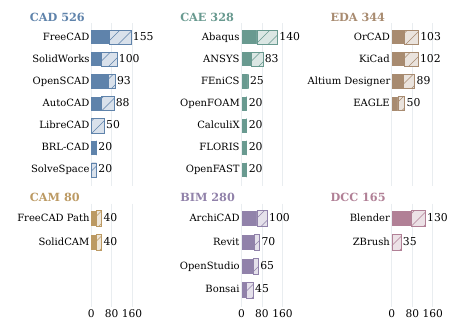}
        \caption{\textbf{Software coverage.}}
        \label{fig:distribution-software}
    \end{subfigure}\hfill
    \begin{subfigure}[t]{0.385\linewidth}
        \centering
        \includegraphics[width=\linewidth]{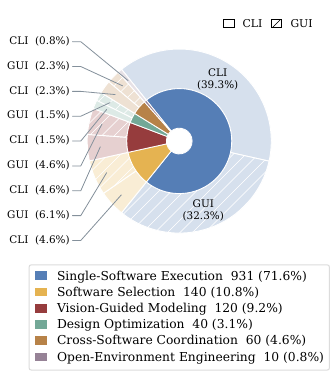}
        \caption{\textbf{Task types and interfaces.}}
        \label{fig:distribution-types}
    \end{subfigure}
    \caption{\textbf{Dataset composition.} \benchmark{} contains 1,301 tasks across six engineering domains and 26 software applications.}
    \label{fig:task-distribution}
\end{figure}

We construct \benchmark{} through an expert-led, AI-assisted pipeline (Figure~\ref{fig:dataset-construction}). Domain experts select the target software and collect real-world tasks from user guides, software forums, textbooks, and other engineering resources. Each task is then built in three parts. Experts write the task instruction, and AI standardizes its structure and wording. The task instance contains the initial files and a ground-truth reference solution, which experts design and construct and AI then refines. For evaluation, experts specify the acceptance criteria, and AI implements them as a task-specific verifier. During the AI-assisted steps, experts review the specifications and executions of every task for correctness, completeness, ambiguity, and evaluability. Before a task enters the benchmark, it is executed and reviewed in its designated software environment (Appendix~\ref{app:construction}).

\benchmark{} contains 1,301 tasks across six engineering domains and 26 software applications. The benchmark includes 691 tasks using the CLI interface and 610 tasks using the GUI interface (Figure~\ref{fig:distribution-types}). Figure~\ref{fig:distribution-software} reports software coverage. The counts add up to 1,723, more than the 1,301 tasks, because some tasks involve several applications, such as Cross-Software Coordination and Software Selection tasks. This view reflects the breadth of the engineering ecosystem covered by \benchmark{}.

\subsection{Artifact-Centric Evaluation}
\label{sec:verification}

To verify the artifacts that agents produce, \benchmark{} evaluates the submission $\mathcal{Y}_T$ using programmatic verifiers. Each verifier reopens the submitted files, extracts the properties required by the task, and checks them against structural requirements or tolerance-bounded numerical criteria. The checks cover geometric dimensions and topology, simulation quantities, schematic connectivity, building-model entities and relationships, and 3D scene structure. For Cross-Software Coordination tasks, verifiers additionally inspect intermediate artifacts and consistency across workflow stages. Missing, invalid, or unreadable files fail the corresponding criteria.

All task types except Design Optimization form the binary set $\mathcal{T}_{\mathrm{bin}}$. Task $i$ has $K_i$ acceptance criteria, each an indicator $\phi_{ik}(\mathcal{Y}_T)\in\{0,1\}$, and the task succeeds only when all are satisfied:
\begin{equation}
    S_i(\mathcal{Y}_T)=\prod_{k=1}^{K_i}\phi_{ik}(\mathcal{Y}_T).
    \label{eq:task-success}
\end{equation}
Design Optimization tasks form the set $\mathcal{T}_{\mathrm{quant}}$ and are evaluated in two stages. A design is feasible only if it satisfies all $L_i$ engineering constraints $\psi_{il}$, and only feasible designs receive a task-specific quality score $q_i\in[0,1]$ (Appendix~\ref{app:scoring}):
\begin{equation}
    % \vspace{-0.1cm}
    F_i(\mathcal{Y}_T)=\prod_{l=1}^{L_i}\psi_{il}(\mathcal{Y}_T),
    \qquad
    Q_i(\mathcal{Y}_T)=
    \begin{cases}
    q_i(\mathcal{Y}_T), & F_i(\mathcal{Y}_T)=1,\\
    0, & F_i(\mathcal{Y}_T)=0.
    \end{cases}
    \label{eq:quantitative-outcome}
    % \vspace{-0.1cm}
\end{equation}

Verifiers are tested on the reference solutions, alternative valid solutions, and intentionally flawed submissions to confirm that they accept valid solutions and reject task-relevant violations. Detailed checker implementations, numerical tolerances, and validation results are provided in Appendix~\ref{app:checks} and Appendix~\ref{app:validation}.
% !TEX root = ../main.tex
\section{Experiments}
\label{sec:experiments}

In this section, we present the experimental setup and the main results of seven frontier models on \benchmark{} (Sections~\ref{sec:setup} and~\ref{sec:main-results}), followed by ablation studies on observation design and interaction history (Section~\ref{sec:ablation}) and a failure analysis (Section~\ref{sec:failures}).

\subsection{Experimental Setup}
\label{sec:setup}

\noindent \textbf{Models.}
We evaluate leading frontier models from major providers: five proprietary models (GPT-5.6 Sol~\citep{openai2026gpt56sol}, Claude Opus 5~\citep{anthropic2026opus5}, Gemini 3.7 Flash~\citep{google2026gemini37flash}, Qwen3.8 Max~\citep{qwen2026qwen38max}, and Qwen3.8 Flash~\citep{qwen2026qwen38next}) and two open-source models (Kimi K3~\citep{moonshot2026kimik3} and DeepSeek V4.1 Flash~\citep{deepseek2026v41flash}). To measure the best performance of each model, all models run at their highest supported reasoning effort. 

\noindent \textbf{Task selection.}
To control evaluation cost, we evaluate all seven models on the same stratified subset of 306 tasks (Table~\ref{tab:subset}). We divide all tasks except Open-Environment Engineering into 72 strata by software, task type, and interface, and randomly select four or five tasks from each stratum, so every combination has a similar weight. Open-Environment Engineering tasks name no software and number only ten, so we include all of them. The resulting subset covers all six engineering domains.

\noindent \textbf{Execution.}
Our execution harness builds on the OSWorld environment~\citep{xie2024osworld}. Each episode starts from a clean Windows or Ubuntu virtual-machine snapshot with task-specific inputs and software states (Appendix~\ref{app:environments}). GUI agents act through PyAutoGUI on $1920\times1080$ screenshots, CLI agents issue terminal commands and scripts, and an image-reading tool lets agents view image files. Vision-Guided Modeling tasks give the reference drawings in the task message. Agents retain the most recent 15 turns of interaction history, which keeps the observations and model outputs without summarization. Episodes are limited to 200 turns for GUI tasks and 100 for CLI tasks, and to five hours of wall-clock time. Longer workflows receive larger budgets: Cross-Software Coordination tasks allow 300 GUI or 150 CLI turns, and Open-Environment Engineering tasks allow 150 turns.

\noindent \textbf{Metrics.}
We report \textbf{EngiScore}, which aggregates binary task success and the quality scores of Design Optimization tasks. For an evaluated task set $\mathcal{D}$,
\begin{equation}
\mathrm{EngiScore}(\mathcal{D}) = \frac{100}{|\mathcal{D}|}
\left(
\smashoperator[r]{\sum_{i\in\mathcal{D}\cap\mathcal{T}_{\mathrm{bin}}}} S_i
+
\smashoperator[r]{\sum_{i\in\mathcal{D}\cap\mathcal{T}_{\mathrm{quant}}}} Q_i
\right),
\label{eq:engiscore}
\end{equation}
where $S_i$ and $Q_i$ are defined in Equations~\eqref{eq:task-success} and~\eqref{eq:quantitative-outcome}. EngiScore ranges from 0 to 100 and weights every task equally. On task sets without Design Optimization tasks, it equals the success rate. We also report three execution metrics: turns per task, output tokens per turn, and API cost per task. A turn is one model call, in which the agent receives the current observation and returns its next actions.

\subsection{Main Results}
\label{sec:main-results}

\noindent \textbf{Overall performance.}
Table~\ref{tab:main-results} reports the main results. Claude Opus 5 achieves the highest overall EngiScore at 44.7, followed by GPT-5.6 Sol at 38.6. Even so, absolute scores remain low, and all seven models receive zero scores on 129 of the 306 tasks (Appendix~\ref{app:all-zero}). Performance also depends strongly on the interface. Every model scores lower on GUI tasks than on CLI tasks, and the failure analysis shows that most GUI failures end at the turn limit. Claude Opus 5 and GPT-5.6 Sol score similarly on CLI tasks (48.6 and 47.9), so the lead of Claude Opus 5 comes from GUI tasks (40.5 versus 28.6). The difference between interfaces is largest for DeepSeek V4.1 Flash, which obtains the second-highest CLI score (48.3) but succeeds on only three GUI tasks, so strong command-line performance does not translate directly to visual interaction.

\begin{table}[!ht]
\centering
\caption{\textbf{Main results on \benchmark{}.} Best and second-best values are \textbf{bold} and \underline{underlined}.}
\label{tab:main-results}
\begingroup
\setlength{\tabcolsep}{2.6pt}
\renewcommand{\arraystretch}{1.05}
\small
\resizebox{\textwidth}{!}{%
\begin{tabular}{@{}l*{12}{c}@{}}
\toprule
& \multicolumn{1}{c}{\textbf{Overall}} & \multicolumn{2}{c}{\textbf{Interface}} & \multicolumn{6}{c}{\textbf{Task type}} & \multicolumn{3}{c}{\textbf{Execution}} \\
\cmidrule(lr){2-2}\cmidrule(lr){3-4}\cmidrule(lr){5-10}\cmidrule(l){11-13}
\textbf{Model} & EngiScore & CLI & GUI & Single & Select. & Vision & Optim. & Cross & Open & Turns & Tokens & Cost \\
 & {\footnotesize (306)} & {\footnotesize (158)} & {\footnotesize (148)} & {\footnotesize (175)} & {\footnotesize (28)} & {\footnotesize (36)} & {\footnotesize (33)} & {\footnotesize (24)} & {\footnotesize (10)} & {\footnotesize (/task)} & {\footnotesize (K/turn)} & {\footnotesize (\$/task)} \\
\midrule
\modellogo{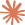}Claude Opus 5 & \textbf{44.7} & \textbf{48.6} & \textbf{40.5} & \textbf{50.3} & \textbf{25.0} & \textbf{66.7} & \underline{32.8} & \textbf{8.3} & 50.0 & \underline{69.5} & 2.4 & 18.78 \\
\modellogo{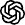}GPT-5.6 Sol & \underline{38.6} & 47.9 & \underline{28.6} & \underline{44.6} & \underline{17.9} & 50.0 & \textbf{36.4} & \underline{4.2} & 40.0 & \textbf{54.2} & 1.0 & 9.57 \\
\modellogo{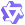}Qwen3.8 Max & 26.3 & 35.3 & 16.6 & 26.3 & 7.1 & \underline{52.8} & 22.3 & 0.0 & \underline{60.0} & 107.3 & 2.7 & 6.44 \\
\modellogo{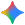}Gemini 3.7 Flash & 26.1 & 37.2 & 14.2 & 30.9 & 3.6 & 30.6 & 20.6 & 0.0 & \textbf{70.0} & 110.0 & \textbf{0.7} & 2.08 \\
\modellogo{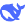}DeepSeek V4.1 Flash & 25.9 & \underline{48.3} & 2.0 & 30.3 & 7.1 & 30.6 & 16.2 & \textbf{8.3} & \underline{60.0} & 115.3 & 9.8 & \underline{0.80} \\
\modellogo{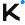}Kimi K3 & 24.3 & 31.3 & 16.7 & 30.3 & 3.6 & 27.8 & 21.9 & 0.0 & 30.0 & 92.1 & \underline{0.9} & 5.85 \\
\modellogo{qwen}Qwen3.8 Flash & 16.7 & 31.5 & 1.0 & 20.0 & 7.1 & 19.4 & 9.9 & \underline{4.2} & 30.0 & 119.6 & 1.3 & \textbf{0.44} \\
\bottomrule
\end{tabular}
}
\endgroup
\end{table}

\begin{figure}[!ht]
\centering
\begin{minipage}[b]{0.38\linewidth}
\centering
\includegraphics[width=\linewidth]{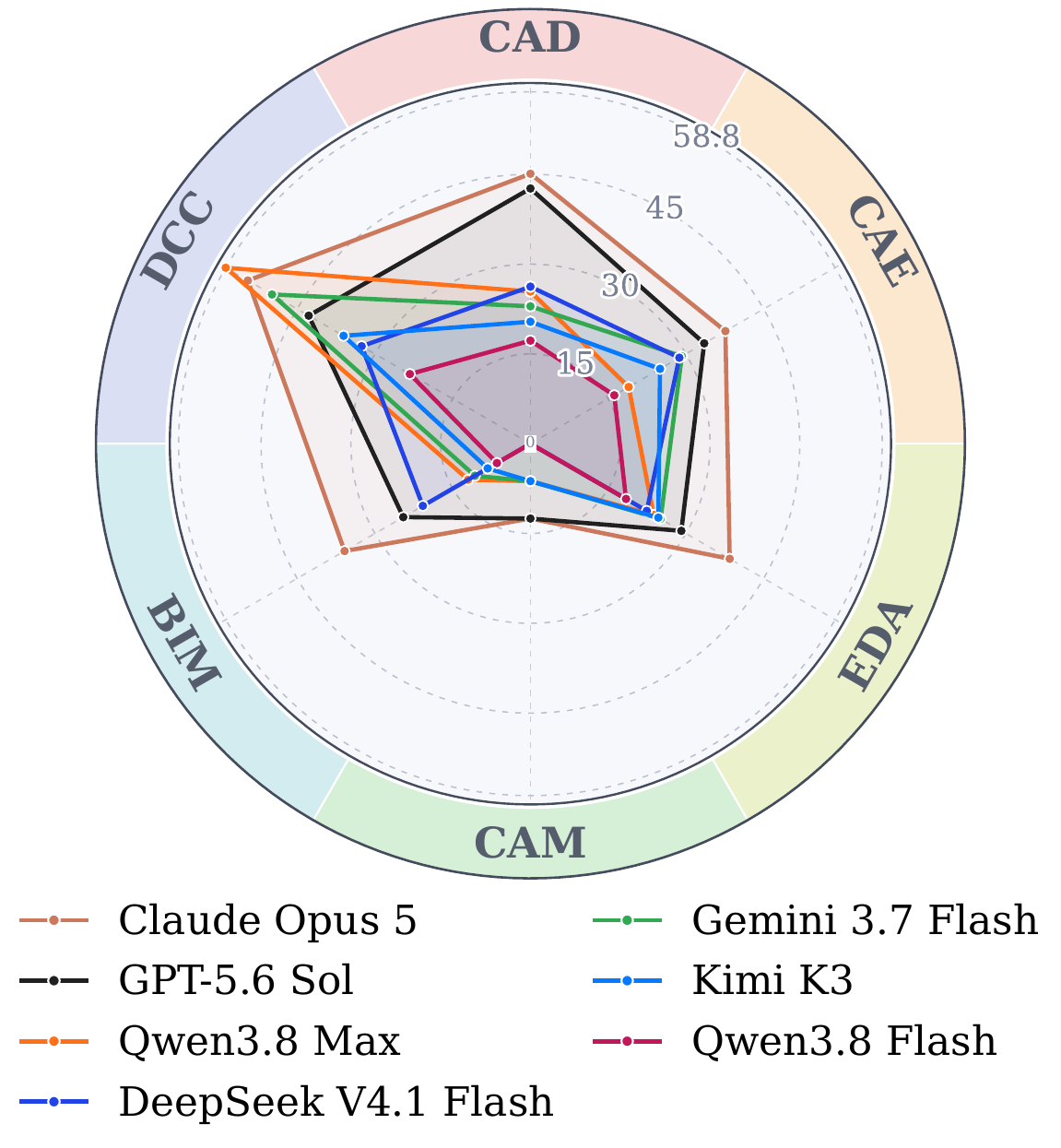}
\captionof{figure}{\textbf{Performance by domain.}}
\label{fig:domain-radar}
\end{minipage}\hfill
\begin{minipage}[b]{0.62\linewidth}
\centering
% Subfigures take their number from the figure counter, which \captionof below steps only afterwards.
\addtocounter{figure}{1}
\begin{subfigure}[b]{0.49\linewidth}
    \centering
    \includegraphics[width=\linewidth]{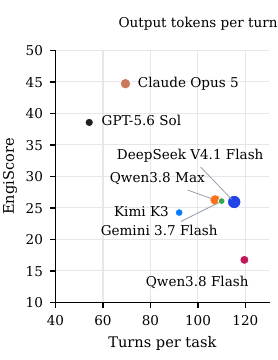}
    \caption{\textbf{Score vs. turns.}}
    \label{fig:score-turns}
\end{subfigure}\hfill
\begin{subfigure}[b]{0.49\linewidth}
    \centering
    \includegraphics[width=\linewidth]{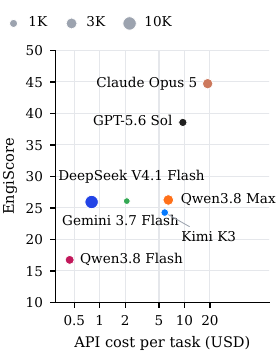}
    \caption{\textbf{Score vs. cost.}}
    \label{fig:score-cost}
\end{subfigure}
\addtocounter{figure}{-1}
\captionof{figure}{\textbf{Score vs. turns and cost.}}
\label{fig:model-comparison}
\end{minipage}
\end{figure}

\begin{figure}[!ht]
\centering
\begin{minipage}[t]{0.35\linewidth}
\centering
\includegraphics[width=\linewidth]{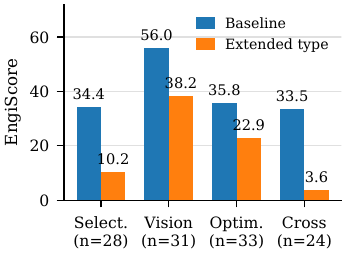}
\captionof{figure}{\textbf{EngiScore of each extended task type.} Baselines are Single-Software Execution tasks of the same model, software, and interface.}
\label{fig:capability-gap}
\end{minipage}\hfill
\begin{minipage}[t]{0.62\linewidth}
\centering
\includegraphics[width=\linewidth]{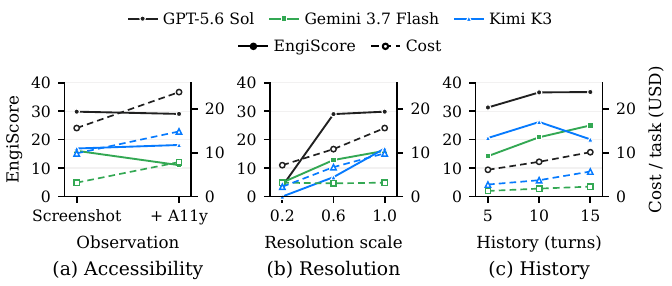}
\captionof{figure}{\textbf{Ablation on observation and history.} Accessibility and resolution use GUI tasks; history includes both interfaces. A11y adds an accessibility tree to the screenshot. Resolution scales are relative to $1920\times1080$.}
\label{fig:observation-history}
\end{minipage}
\end{figure}

\noindent \textbf{Task types.}
Performance drops sharply when a task requires coordination across software. Claude Opus 5 reaches an EngiScore of 50.3 on Single-Software Execution tasks and 66.7 on Vision-Guided Modeling tasks, but only 6 of the 168 Cross-Software Coordination runs succeed across all models. Software Selection tasks are also difficult, and even the strongest model solves only one quarter of them. These raw scores, however, mix the effect of each capability with that of the software: Vision-Guided Modeling tasks score higher than Single-Software Execution tasks because they use software on which agents perform well. To separate the two effects, we compare each extended task type, that is, each of the five types beyond Single-Software Execution, with Single-Software Execution tasks of the same model, software, and interface. Open-Environment Engineering is left out because its tasks name no software. Figure~\ref{fig:capability-gap} shows that each of the other four capabilities lowers performance. Agents keep 68\% of their single-software score when they read drawings, 64\% when they optimize a design, 30\% when they select software, and 11\% when they coordinate several applications. These results point to a central limitation of agents: they can complete tasks within one application but break down when a task requires selecting the right software, transferring intermediate artifacts, or preserving dependencies across a multi-stage workflow.

\noindent \textbf{Design Optimization.}
Most Design Optimization runs score zero. Of the 231 runs, 169 (73\%) receive a zero score, 53 a partial score, and 9 a full score (Appendix~\ref{app:optimization}). Only Claude Opus 5 and GPT-5.6 Sol reach a full score, and partial scores concentrate on two task families: KiCad layout and ZBrush surface tasks account for 38 of the 53. A design must pass the feasibility checks to receive a nonzero score, and the quality score then ranks the designs that pass. On ZBrush surface tasks, for example, invalid geometry sets the score to zero regardless of how closely the surface approximates the target.

\noindent \textbf{Open-Environment Engineering.}
Each Open-Environment Engineering task keeps the engineering target of a Cross-Software Coordination task but drops the prescribed software, stage order, and intermediate handoffs, so only the final design is checked. On the four tasks evaluated in both versions, all 28 Cross-Software Coordination runs fail, whereas 13 of the 28 Open-Environment Engineering runs succeed (Appendix~\ref{app:paired}). Thus, agents can reach these engineering targets, and the handoffs between applications are the main obstacle in Cross-Software Coordination tasks. Over all ten tasks, Open-Environment Engineering reaches an EngiScore of 48.6, close to the 46.3 on CLI Single-Software Execution tasks. The two types measure different abilities: operating a given application, and finding software one handles well to solve a problem.

\noindent \textbf{Engineering domains.}
Figure~\ref{fig:domain-radar} compares the models across the six engineering domains and shows substantial specialization. Every model scores highest in DCC and lowest in CAM, where no model exceeds 12.5 on the 16 CAM tasks and two models score zero. Claude Opus 5 leads in CAD, CAE, EDA, and BIM and ties GPT-5.6 Sol in CAM, whereas Qwen3.8 Max, third overall, performs best in DCC. The gap between the leading models and the others depends on the domain. In CAD, Claude Opus 5 and GPT-5.6 Sol score 45.1 and 42.6, while no other model exceeds 26.2. The gap narrows in CAE, where Gemini 3.7 Flash and DeepSeek V4.1 Flash reach 29.2 and 28.7, against 33.5 for GPT-5.6 Sol. BIM separates Claude Opus 5 from all other models, including GPT-5.6 Sol (35.8 versus 24.5).

\noindent \textbf{Execution cost.}
Figure~\ref{fig:model-comparison} plots EngiScore against turns and API cost per task. GPT-5.6 Sol uses the fewest turns per task, followed by Claude Opus 5. DeepSeek V4.1 Flash and Gemini 3.7 Flash achieve similar scores of approximately 26, but the former costs less than half as much. The difference comes from per-token pricing, since DeepSeek V4.1 Flash generates about 15 times as many output tokens per turn as Gemini 3.7 Flash and still costs less. Claude Opus 5 reaches the highest score at nearly twice the API cost of GPT-5.6 Sol. Across all models, failed runs account for most of the turns. In every task type except Open-Environment Engineering, failed runs use 1.7 to 4.0 times as many turns as successful runs (Appendix~\ref{app:execution-by-type}).

\subsection{Ablation and Analysis}
\label{sec:ablation}

We run five ablations with GPT-5.6 Sol, Gemini 3.7 Flash, and Kimi K3 on task subsets to study how observations and interaction history affect agent performance. \emph{Initial-image presentation} gives the reference drawings of Vision-Guided Modeling tasks either as files that the agent must open or directly in the task message. \emph{Runtime image access} turns the image-reading tool of CLI agents on or off. The tool lets agents view reference drawings and renderings of their own intermediate artifacts, and we test it on tasks with and without a reference drawing. \emph{Accessibility information} adds an accessibility tree to each GUI screenshot. \emph{Screenshot resolution} scales GUI screenshots to 0.2, 0.6, or 1.0 times the native $1920\times1080$ resolution. \emph{History length} keeps the most recent 5, 10, or 15 turns of interaction history. Appendix~\ref{app:ablations} details the task subsets and task-level gains and regressions.

\begin{table}[!ht]
\centering
\begin{minipage}[b]{0.48\linewidth}
\centering
\caption{\textbf{Ablation on initial-image presentation.} EngiScore with reference drawings as files in the environment (Env.) or in the task message (Msg.).}
% \vspace{-0.1cm}
\label{tab:initial-image}
\begingroup
\setlength{\tabcolsep}{2pt}
\renewcommand{\arraystretch}{1.0}
\small
\begin{tabular}{@{}lcccc@{}}
\toprule
& \multicolumn{2}{c}{\textbf{CLI}} & \multicolumn{2}{c}{\textbf{GUI}} \\
\cmidrule(lr){2-3}\cmidrule(l){4-5}
\textbf{Model} & Env. & Msg. & Env. & Msg. \\
\midrule
\modellogo{openai}GPT-5.6 Sol & 63.3 & 68.3 & 15.0 & 31.7 \\
\modellogo{gemini}Gemini 3.7 Flash & 38.3 & 40.0 & 3.3 & 10.0 \\
\modellogo{kimi}Kimi K3 & 40.0 & 36.7 & 6.7 & 10.0 \\
\bottomrule
\end{tabular}
\endgroup
\end{minipage}
\hfill
\begin{minipage}[b]{0.48\linewidth}
\centering
\caption{\textbf{Ablation on runtime image access in CLI.} EngiScore with the image-reading tool disabled (Off) or enabled (On), on tasks with and without an initial reference drawing.}
% \vspace{-0.1cm}
\label{tab:cli-visual}
\begingroup
\setlength{\tabcolsep}{2pt}
\renewcommand{\arraystretch}{1.0}
\small
\begin{tabular}{@{}lcccc@{}}
\toprule
& \multicolumn{2}{c}{\textbf{w/ Init. Image}} & \multicolumn{2}{c}{\textbf{w/o Init. Image}} \\
\cmidrule(lr){2-3}\cmidrule(l){4-5}
\textbf{Model} & Off & On & Off & On \\
\midrule
\modellogo{openai}GPT-5.6 Sol & 66.7 & 68.3 & 48.1 & 41.5 \\
\modellogo{gemini}Gemini 3.7 Flash & 36.7 & 40.0 & 33.3 & 36.6 \\
\modellogo{kimi}Kimi K3 & 41.7 & 36.7 & 31.5 & 26.5 \\
\bottomrule
\end{tabular}
\endgroup
\end{minipage}
% \vspace{-0.2cm}
\end{table}

\noindent \textbf{Initial-image presentation.}
Presenting reference drawings directly in the task message improves GUI performance for all three models, with GPT-5.6 Sol increasing from 15.0 to 31.7 (Table~\ref{tab:initial-image}). GUI agents otherwise have to find and open the drawing through the interface, and the message removes this retrieval step. On the CLI interface, where the image-reading tool already gives direct access to the drawings, the gain is smaller, and Kimi K3 even declines. However, five GUI tasks that GPT-5.6 Sol previously solved now fail. Easier access to drawings thus helps agents find task information but does not ensure that they use it correctly.

\noindent \textbf{Runtime image access.}
Enabling the image-reading tool improves Gemini 3.7 Flash and lowers Kimi K3, both with and without a reference drawing (Table~\ref{tab:cli-visual}). GPT-5.6 Sol improves when a reference drawing is given and declines when it is not. A rendered view lets an agent compare the features of its model with the drawing, but it cannot confirm that their depths and angles are exact, and it can also trigger unnecessary revisions. Whether the tool helps therefore depends on how each model uses the images, and rendered views cannot replace measuring the artifact.

\noindent \textbf{Accessibility information.}
Adding an accessibility tree improves Kimi K3, barely changes GPT-5.6 Sol, and lowers Gemini 3.7 Flash (Figure~\ref{fig:observation-history}a). All three models use fewer turns, but API cost rises by 51\% to 148\%. The tree helps agents locate interface controls, but engineering operations still depend on selecting the right body, placement, and feature direction. It thus shortens the interaction but does not consistently improve the resulting artifacts.

\noindent \textbf{Screenshot resolution.}
The native $1920\times1080$ resolution produces the highest score for all three models (Figure~\ref{fig:observation-history}b). At $0.6\times$, GPT-5.6 Sol keeps most of its score at about 30\% lower cost, while Kimi K3 loses much more. Moving from $0.6\times$ to $1.0\times$ solves new tasks for Kimi K3 without losing any, while the gains and losses of GPT-5.6 Sol nearly cancel. Finer detail helps agents distinguish small features, such as the pads and apertures of a PCB footprint. A lower resolution is thus a reasonable way to cut cost for GPT-5.6 Sol but not for Kimi K3.

\noindent \textbf{Interaction history.}
The best history length depends on the model (Figure~\ref{fig:observation-history}c). Gemini 3.7 Flash improves as the history grows from 5 to 15 turns, GPT-5.6 Sol changes little beyond 10 turns, and Kimi K3 performs best at 10. For Kimi K3, extending the history from 10 to 15 turns lowers EngiScore and raises cost by about half, mainly because 11 tasks lose their nonzero score. A longer history is therefore not a free improvement, and its length should be set for each model.
% \vspace{-0.1cm}
\FloatBarrier
\subsection{Failure Analysis}
% \vspace{-0.1cm}
\label{sec:failures}

\begin{figure}[!htb]
\centering
\begin{subfigure}[b]{0.52\linewidth}
    \centering
    \includegraphics[width=\linewidth]{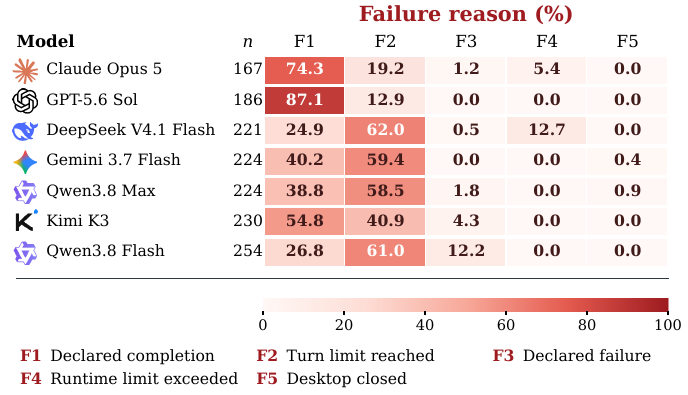}
    \caption{\textbf{Failure modes by model.}}
    \label{fig:failure-reasons}
\end{subfigure}\hfill
\begin{subfigure}[b]{0.48\linewidth}
    \centering
    \includegraphics[width=\linewidth]{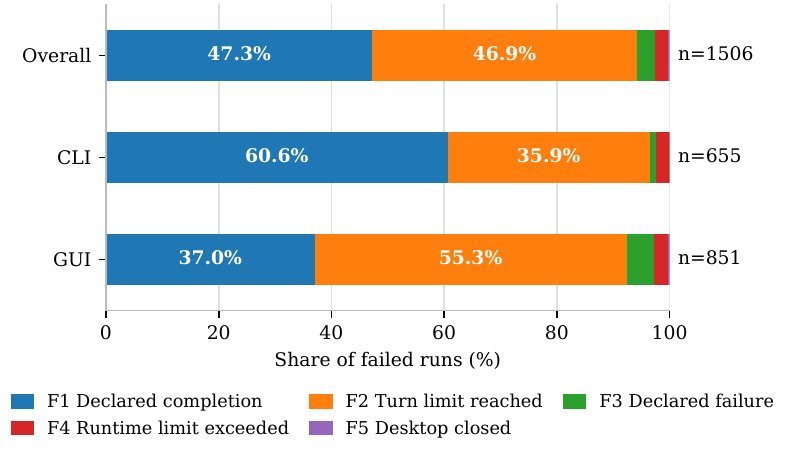}
    \caption{\textbf{Failure modes by interface.}}
    \label{fig:interface-failure-breakdown}
\end{subfigure}
\caption{\textbf{Failure modes in the main evaluation.} Failed runs are grouped by terminal outcome.}
\label{fig:failure-analysis}
\end{figure}

We treat runs with a zero score as failures and classify them by their terminal outcome: declared completion, reaching the turn limit, declared failure, runtime-limit violation, and unexpected desktop closure or restart. Figure~\ref{fig:failure-analysis} breaks down the failures by model and by interface. Two outcomes account for 94\% of failures in the main evaluation. Agents either declare completion without detecting the errors in their artifacts or run out of turns before the artifact is complete. The first reflects weak self-verification, and the second reflects slow progress over long interactions.

\noindent \textbf{Completion without verification.}
The two strongest models fail mainly by declaring completion too early. Completion declarations end 87\% of the failures of GPT-5.6 Sol and 74\% of those of Claude Opus 5, compared with 40\% for Gemini 3.7 Flash. These runs mostly fail on incorrect geometry or broken consistency across workflow stages (Appendix~\ref{app:violations}). For example, one run cuts a microfluidic channel from the wrong face, and another assigns a mounting hole a role that does not match the source PCB, while its final report states that all requirements are met. Reliable engineering agents therefore need to verify their artifacts against the task requirements before declaring completion.

\noindent \textbf{Running out of turns on GUI tasks.}
Weaker models and GUI agents fail mainly by reaching the turn limit. It is the most common failure outcome for Gemini 3.7 Flash, Qwen3.8 Max, Qwen3.8 Flash, and DeepSeek V4.1 Flash, and accounts for 55\% of GUI failures but only 36\% of CLI failures (Figure~\ref{fig:interface-failure-breakdown}). GUI agents reach every operation through screenshots and mouse or keyboard actions, so a modeling workflow takes far more turns than the equivalent CLI script. Declared failures, runtime-limit violations, and desktop closures are much less frequent, and all runtime-limit violations come from DeepSeek V4.1 Flash and Claude Opus 5, whose runs reach the five-hour limit (Table~\ref{tab:failure-main}). Efficient GUI operation is thus as important for engineering agents as engineering knowledge.

\noindent \textbf{Errors at software handoffs.}
Handoffs between applications are a separate source of failure. They cause two of the four Cross-Software Coordination failures examined in Appendix~\ref{app:cross-failures}. A DXF import loses the model scale, and a CSV of board parameters uses the wrong column name, so the downstream applications and checks cannot use the transferred data. The other two runs fail inside one application, through a load defined as a pressure where the task requires a force and a GUI action that deletes the drawing content. Each handoff adds a point at which a correct artifact becomes unusable for the next application. This is consistent with Cross-Software Coordination tasks keeping only 11\% of the single-software score, and with the paired Open-Environment Engineering tasks, which the same models solve once the prescribed handoffs are removed.

% !TEX root = ../main.tex
% \vspace{-0.1cm}
\FloatBarrier
\section{Conclusion}
% \vspace{-0.1cm}
\label{sec:conclusion}
We introduced \benchmark{}, a benchmark that evaluates agents on the capabilities that real engineering projects require, in 26 professional software applications across six engineering domains. Its 1,301 expert-curated tasks range from operating one application to solving open-ended engineering problems, and the artifact-centric evaluation judges the engineering validity of what agents deliver. Our experiments show that frontier agents can complete many tasks within one application but rarely carry a design across several, which makes cross-software coordination the central open problem. We hope \benchmark{} serves as a common testbed for building agents that take on real engineering projects in industry.

% \section*{Limitations}
% Our evaluation covers a 306-task subset of \benchmark{} with a single run per model, so it does not measure run-to-run variance. The verifiers check the engineering properties specified by each task and do not certify designs for real-world use. The Open-Environment Engineering tasks are all derived from one cross-software workflow and use only the CLI interface.

% \clearpage
\bibliographystyle{plainnat}
\bibliography{references}

\clearpage
\beginappendix
\counterwithin{figure}{section}
\counterwithin{table}{section}
\counterwithin{equation}{section}
\makeatletter
\setlength{\@fptop}{0pt}
\setlength{\@fpsep}{12pt}
\renewcommand{\topfraction}{0.9}
\renewcommand{\bottomfraction}{0.8}
\renewcommand{\textfraction}{0.07}
\renewcommand{\floatpagefraction}{0.85}
\setcounter{topnumber}{4}
\setcounter{bottomnumber}{3}
\setcounter{totalnumber}{6}
\makeatother
% !TEX root = ../main.tex
\FloatBarrier
\section{Benchmark Construction}
\label{app:construction}

\benchmark{} is built through an expert-led, AI-assisted pipeline (Figure~\ref{fig:dataset-construction}). Domain experts define task objectives, engineering artifacts, and acceptance criteria. AI assists in standardizing specifications, preparing artifacts, and implementing verifiers. Experts review the specifications and executions of every task, and a task enters the benchmark only after its instruction, initial state, reference solution, and verifier pass a final joint review.

\subsection{Task Specification}
\label{app:task-spec}

For each of the 26 software applications, experts select scenarios that represent its characteristic capabilities and workflows, drawing on official documentation, user guides, tutorials, technical references, and engineering textbooks. Each scenario is written as a result-oriented specification that states the engineering objective, available inputs, required deliverables, and constraints. AI standardizes the structure and wording so that tasks in different software applications follow one format. When a task starts from existing artifacts, properties already contained in those artifacts are not restated, and the instruction focuses on the intended outcome and the deliverables.

\subsection{Task Instances and Reference Solutions}
\label{app:instances}

Experts prepare the initial engineering state in the designated software and build a reference solution, with AI assistance during artifact preparation. Depending on the domain, an instance may contain CAD geometry, PCB designs, simulation models, BIM models, manufacturing files, structured parameters, or 3D scenes. The reference solution confirms that the task is solvable and provides the engineering quantities used to calibrate the verifier. It is not the only accepted answer: any submission that meets the engineering requirements passes. In Cross-Software Coordination tasks, intermediate artifacts carry the dependencies from one stage to the next. In Open-Environment Engineering tasks, the requirements and deliverables are fixed, while the choice of software and execution order is left to the agent.

\subsection{Evaluated Subset}
\label{app:subset}

Table~\ref{tab:subset} lists the number of tasks of each type and interface in the full benchmark and in the 306-task subset used in the main evaluation.

\begin{table}[!ht]
\centering
\appendixtablestyle
\caption{\textbf{Task counts in the full benchmark and the evaluated subset.}}
\label{tab:subset}
\begin{tabular}{@{}lrrrr@{}}
\toprule
 & \multicolumn{2}{c}{\textbf{Full benchmark}} & \multicolumn{2}{c}{\textbf{Evaluated subset}} \\
\cmidrule(lr){2-3}\cmidrule(l){4-5}
\textbf{Task type} & CLI & GUI & CLI & GUI \\
\midrule
Single-Software Execution & 511 & 420 & 92 & 83 \\
Software Selection & 60 & 80 & 12 & 16 \\
Vision-Guided Modeling & 60 & 60 & 16 & 20 \\
Design Optimization & 20 & 20 & 16 & 17 \\
Cross-Software Coordination & 30 & 30 & 12 & 12 \\
Open-Environment Engineering & 10 & -- & 10 & -- \\
\midrule
Total & 691 & 610 & 158 & 148 \\
\bottomrule
\end{tabular}
\end{table}

\subsection{Software Environments}
\label{app:environments}

Each task runs in an isolated Windows or Ubuntu virtual machine, which has the required software installed except in Open-Environment Engineering tasks. \benchmark{} provides 37 environments (Table~\ref{tab:environments}): 26 software-specific environments, in which FreeCAD and FreeCAD Path count as separate software applications; 10 composite environments for Cross-Software Coordination and Software Selection tasks; and one environment for Open-Environment Engineering tasks. Before each episode, the harness restores a clean snapshot, copies the task inputs, and opens the relevant documents. In Cross-Software Coordination tasks, all participating applications run in the same virtual machine and read and write the same working directory. Table~\ref{tab:environments} lists versions for the software-specific environments.

\begin{table}[!ht]
\centering
\small
\setlength{\tabcolsep}{4pt}
\renewcommand{\arraystretch}{1.0}
\caption{\textbf{Software environments.}}
\label{tab:environments}
\begin{tabularx}{\linewidth}{@{}l>{\raggedright\arraybackslash}Xlll@{}}
\toprule
\textbf{Domain / type} & \textbf{Software} & \textbf{Version} & \textbf{OS} & \textbf{Interface} \\
\midrule
\multicolumn{5}{l}{\textbf{Software-specific environments (26)}} \\
\midrule
CAD & FreeCAD & 0.21.2 & Ubuntu & CLI/GUI \\
    & AutoCAD & 2024 & Windows & CLI/GUI \\
    & SolidWorks & 2025 & Windows & CLI/GUI \\
    & BRL-CAD & 7.32.2 & Ubuntu & CLI \\
    & OpenSCAD & 2021.01 & Ubuntu & CLI/GUI \\
    & LibreCAD & 2.2.0.2 & Ubuntu & GUI \\
    & SolveSpace & 3.1 & Ubuntu & GUI \\
\midrule
CAE & ANSYS & 2026 R1 & Windows & CLI/GUI \\
    & Abaqus & 2025 (Learning Edition) & Windows & CLI/GUI \\
    & OpenFOAM & 11 & Ubuntu & CLI \\
    & FEniCS (DOLFIN) & 2019.1.0 & Ubuntu & CLI \\
    & CalculiX & 2.21 & Ubuntu & CLI \\
    & FLORIS & 4.6.4 & Ubuntu & CLI \\
    & OpenFAST & 5.0.0 & Ubuntu & CLI \\
\midrule
EDA & Cadence OrCAD & 24.1 & Windows & CLI/GUI \\
    & EAGLE & 7.7.0 & Ubuntu & CLI/GUI \\
    & Altium Designer & 17 & Windows & CLI/GUI \\
    & KiCad & 10.0.2 & Ubuntu & CLI/GUI \\
\midrule
CAM & FreeCAD Path & 0.21.2 & Ubuntu & CLI/GUI \\
    & SolidCAM & 2025 & Windows & CLI/GUI \\
\midrule
BIM & Archicad & 27 & Windows & CLI/GUI \\
    & Bonsai & 0.8.5 & Ubuntu & CLI/GUI \\
    & OpenStudio & 1.11.0 & Ubuntu & CLI/GUI \\
    & Revit & 2025 & Windows & CLI/GUI \\
\midrule
DCC & Blender & 4.2.3 & Ubuntu & CLI/GUI \\
        & ZBrush & 2025 & Windows & GUI \\
\midrule
\multicolumn{5}{l}{\textbf{Composite environments: Cross-Software Coordination tasks (6)}} \\
\midrule
Cross & AutoCAD + SolidWorks + Abaqus & --- & Windows & GUI \\
      & SolidWorks + Abaqus & --- & Windows & GUI \\
      & Archicad + OpenStudio & --- & Windows & CLI \\
      & Revit + Archicad + OpenStudio & --- & Windows & CLI \\
      & LibreCAD + FreeCAD + KiCad + Blender & --- & Ubuntu & GUI \\
      & KiCad + OpenSCAD + FreeCAD + Blender & --- & Ubuntu & CLI \\
\midrule
\multicolumn{5}{l}{\textbf{Composite environments: Software Selection tasks (4)}} \\
\midrule
Selection & ANSYS + Abaqus + AutoCAD & --- & Windows & CLI/GUI \\
          & OrCAD + Altium Designer + OpenSCAD & --- & Windows & CLI/GUI \\
          & Revit + Archicad + Abaqus & --- & Windows & CLI/GUI \\
          & FreeCAD + LibreCAD + KiCad & --- & Ubuntu & GUI \\
\midrule
\multicolumn{5}{l}{\textbf{Open-Environment Engineering environment (1)}} \\
\midrule
Open & Computer without preinstalled engineering software & --- & Ubuntu & CLI \\
\bottomrule
\end{tabularx}
\end{table}

% !TEX root = ../main.tex
\FloatBarrier
\needspace{8\baselineskip}
\section{Evaluation Protocol}
\label{app:evaluation}

\subsection{Artifact Checks}
\label{app:checks}

Acceptance criteria follow directly from the engineering requirements of each task. Experts specify the properties and tolerances that define success, and AI helps translate them into a task-specific verifier. After an episode ends, the verifier runs inside the task virtual machine, reopens the submitted deliverables through the software's native data model or a format-specific parser, and checks the extracted properties (Table~\ref{tab:checks}). The parsers include CadQuery/OCCT for solids, ezdxf for drawings, IfcOpenShell for building models, the FreeCAD and Blender APIs, the Abaqus output-database API for simulation results, and SQLite for EnergyPlus/OpenStudio outputs.

Neutral formats such as STEP, DXF, and IFC are re-imported, so export errors affect the result. Numerical properties use task-specific tolerances, whereas connectivity, entity roles, and required references must match exactly. Equivalent names or representations permitted by a file format are accepted. Missing or unreadable deliverables fail the corresponding criteria, and integrity checks protect designated input files and detect GUI-interface tasks that are completed through scripts instead of the GUI.

For Cross-Software Coordination tasks, verifiers also check intermediate artifacts against the task inputs. In a KiCad--OpenSCAD--FreeCAD--Blender workflow, for example, they match board identifiers and mounting-hole dimensions to the mechanical map (the CSV of board parameters exported from KiCad) and then verify that the enclosure, assembly, and review scene preserve these features. A missing upstream artifact fails every criterion that depends on it.

\begin{table}[!ht]
\centering
\appendixtablestyle
\caption{\textbf{Artifact checks.} Representative properties and acceptance rules across the verifier families. Numerical tolerances are task specific.}
\label{tab:checks}
\begingroup
\begin{tabularx}{\linewidth}{@{}>{\raggedright\arraybackslash}p{0.20\linewidth}>{\raggedright\arraybackslash}X>{\raggedright\arraybackslash}p{0.25\linewidth}@{}}
\toprule
\textbf{Check family} & \textbf{Properties} & \textbf{Acceptance criteria} \\
\midrule
Global geometry & Bounding spans, volume, surface area, and solid validity & Dimensional bounds and topology checks \\
Local geometry & Profiles at specified sections in the task's datum frame & Profile tolerances, e.g., $0.005$--$0.08$\,mm \\
Solid--void occupancy & Material, cavities, through-holes, and bore clearance & Required occupancy; residual material, e.g., $\leq0.1$\,mm$^3$ \\
Engineering assignments & Materials, loads, boundary conditions, and thermal zones & Required assignments and parameter values \\
Simulation outputs & Field quantities and physical responses & Task-specific absolute or relative error bounds \\
Structural relations & Net connectivity, entity roles, and cross-file references & Exact connectivity and semantic consistency \\
\bottomrule
\end{tabularx}
\endgroup
\end{table}

\subsection{Scoring Design Optimization}
\label{app:scoring}

Design Optimization tasks combine hard feasibility constraints $\psi_{il}$ (Equation~\eqref{eq:quantitative-outcome}), such as required geometry, material and load assignments, and limits on physical response, with a design objective $m_i$. When the objective measures improvement over an initial design, the quality score $q_i$ is normalized between a baseline $m_{i,\mathrm{base}}$ and an improved reference $m_{i,\mathrm{ref}}$:
\begin{equation}
q_i(\mathcal{Y}_T)=
\operatorname{clip}_{[0,1]}\!\left(\dfrac{m_i(\mathcal{Y}_T)-m_{i,\mathrm{base}}}{m_{i,\mathrm{ref}}-m_{i,\mathrm{base}}}\right).
\label{eq:quant-score}
\end{equation}
For both minimization and maximization objectives, matching the baseline gives zero and matching or exceeding the reference gives one. Other tasks score design quality directly as $q_i(\mathcal{Y}_T)=\max\bigl(0,\,1-\lambda_i c_i(\mathcal{Y}_T)\bigr)$, where $c_i$ is a task-specific weighted cost, such as PCB layout cost or surface approximation error with penalty terms, and $\lambda_i$ is a fixed scale. Because this score is not normalized by the initial design, a positive score does not necessarily indicate improvement over it. Infeasible outputs receive zero regardless of the objective value. Appendix~\ref{app:optimization} reports raw measurements for representative tasks.

\subsection{Verifier Validation}
\label{app:validation}

Each verifier is validated together with its task instance in the target software. The reference solution must pass, intentionally flawed submissions must be rejected, and for tasks with several valid solutions, alternative solutions must also pass. Expected outcomes are assigned from the task requirements before the verifier runs: task authors label alternative and defective artifacts, and a second annotator reviews these labels independently of the verifier output. Disagreements trigger a review of the specification and a rerun of the affected checks.

Table~\ref{tab:verifier-validation} summarizes the audit, which covers 96 randomly sampled reference solutions. All 96 pass after review and repair. Alternative solutions for the Abaqus/ANSYS Software Selection tasks and for FreeCAD lightweight design are all accepted, while defective submissions, namely empty deliveries and reports that state metrics without delivering artifacts, are all rejected.

\begin{table}[!ht]
\centering
\appendixtablestyle
\caption{\textbf{Verifier validation.} Acceptance outcomes for reference, alternative-valid, and defective artifact populations.}
\label{tab:verifier-validation}
\begingroup
\begin{tabular}{@{}llrrr@{}}
\toprule
\textbf{Test population} & \textbf{Scope} & \textbf{$n$} & \textbf{Accepted} & \textbf{Rejected} \\
\midrule
Reference solutions & Audit group & 96 & 96 & 0 \\
Alternative solver & Abaqus / ANSYS, CLI & 40 & 40 & 0 \\
Alternative solver & Abaqus / ANSYS, GUI & 40 & 40 & 0 \\
Alternative design & FreeCAD lightweight design & 5 & 5 & 0 \\
\midrule
Empty delivery & Software Selection tasks, CLI / GUI & 21 & 0 & 21 \\
Metrics without artifacts & Software Selection tasks, CLI / GUI & 21 & 0 & 21 \\
\bottomrule
\end{tabular}
\endgroup
\end{table}

% !TEX root = ../main.tex
\FloatBarrier
\section{Additional Results}
\label{app:results}

\subsection{Results by Interface}
\label{app:interface}

Table~\ref{tab:main-by-interface} breaks down the main results by interface.

\begin{table}[!ht]
\centering
\appendixtablestyle
\setlength{\tabcolsep}{3pt}
\caption{\textbf{Main results by interface.} Columns follow Table~\ref{tab:main-results}; Open-Environment Engineering tasks use only the CLI interface.}
\label{tab:main-by-interface}
\begin{tabular}{@{}l*{10}{c}@{}}
\toprule
& \multicolumn{1}{c}{\textbf{Overall}} & \multicolumn{6}{c}{\textbf{Task type}} & \multicolumn{3}{c}{\textbf{Execution}} \\
\cmidrule(lr){2-2}\cmidrule(lr){3-8}\cmidrule(l){9-11}
\textbf{Model} & EngiScore & Single & Select. & Vision & Optim. & Cross & Open & Turns & Tokens & Cost \\
\midrule
\multicolumn{11}{c}{\textit{CLI interface}} \\
\midrule
\modellogo{claude}Claude Opus 5 & \textbf{48.6} & \underline{55.4} & \textbf{33.3} & \textbf{75.0} & 17.7 & \textbf{16.7} & 50.0 & \underline{33.5} & 3.3 & 8.96 \\
\modellogo{openai}GPT-5.6 Sol & 47.9 & 53.3 & \textbf{33.3} & \textbf{75.0} & \textbf{35.2} & \underline{8.3} & 40.0 & \textbf{28.1} & 1.4 & 3.71 \\
\modellogo{qwen}Qwen3.8 Max & 35.3 & 38.0 & 0.0 & \underline{68.8} & 23.2 & 0.0 & \underline{60.0} & 57.2 & 2.7 & 2.43 \\
\modellogo{gemini}Gemini 3.7 Flash & 37.2 & 44.6 & 0.0 & 50.0 & 17.7 & 0.0 & \textbf{70.0} & 58.4 & \textbf{0.9} & 1.08 \\
\modellogo{deepseek}DeepSeek V4.1 Flash & \underline{48.3} & \textbf{56.5} & \underline{16.7} & 56.2 & \underline{33.5} & \textbf{16.7} & \underline{60.0} & 35.7 & 15.6 & \underline{0.54} \\
\modellogo{kimi}Kimi K3 & 31.3 & 39.1 & 8.3 & 37.5 & 21.9 & 0.0 & 30.0 & 53.7 & \underline{1.1} & 1.94 \\
\modellogo{qwen}Qwen3.8 Flash & 31.5 & 37.0 & \underline{16.7} & 43.8 & 17.7 & \underline{8.3} & 30.0 & 62.7 & 1.9 & \textbf{0.18} \\
\midrule
\multicolumn{11}{c}{\textit{GUI interface}} \\
\midrule
\modellogo{claude}Claude Opus 5 & \textbf{40.5} & \textbf{44.6} & \textbf{18.8} & \textbf{60.0} & \textbf{47.1} & 0.0 & -- & \underline{107.9} & 1.4 & 29.27 \\
\modellogo{openai}GPT-5.6 Sol & \underline{28.6} & \underline{34.9} & 6.2 & 30.0 & \underline{37.6} & 0.0 & -- & \textbf{82.1} & \underline{0.6} & 15.82 \\
\modellogo{qwen}Qwen3.8 Max & 16.6 & 13.3 & \underline{12.5} & \underline{40.0} & 21.4 & 0.0 & -- & 160.7 & 2.6 & 10.72 \\
\modellogo{gemini}Gemini 3.7 Flash & 14.2 & 15.7 & 6.2 & 15.0 & 23.4 & 0.0 & -- & 165.0 & \textbf{0.4} & 3.16 \\
\modellogo{deepseek}DeepSeek V4.1 Flash & 2.0 & 1.2 & 0.0 & 10.0 & 0.0 & 0.0 & -- & 200.2 & 3.5 & \underline{1.08} \\
\modellogo{kimi}Kimi K3 & 16.7 & 20.5 & 0.0 & 20.0 & 21.9 & 0.0 & -- & 133.1 & \underline{0.6} & 10.01 \\
\modellogo{qwen}Qwen3.8 Flash & 1.0 & 1.2 & 0.0 & 0.0 & 2.5 & 0.0 & -- & 180.2 & 0.7 & \textbf{0.72} \\
\bottomrule
\end{tabular}
\end{table}

\subsection{Execution Metrics by Task Type}
\label{app:execution-by-type}

Cross-Software Coordination tasks require the most turns, time, and cost (Table~\ref{tab:execution-by-type}). Compared with Single-Software Execution tasks, they use 2.2 times the turns, 2.0 times the wall-clock time, and 3.4 times the API cost. Software Selection tasks, which add the choice of software, also use more turns than Single-Software Execution tasks (112.3 versus 85.8) under similar turn limits. Cross-Software Coordination tasks allow more turns, and their failed runs mostly use all of them.

\begin{figure}[!ht]
\centering
\includegraphics[width=\linewidth]{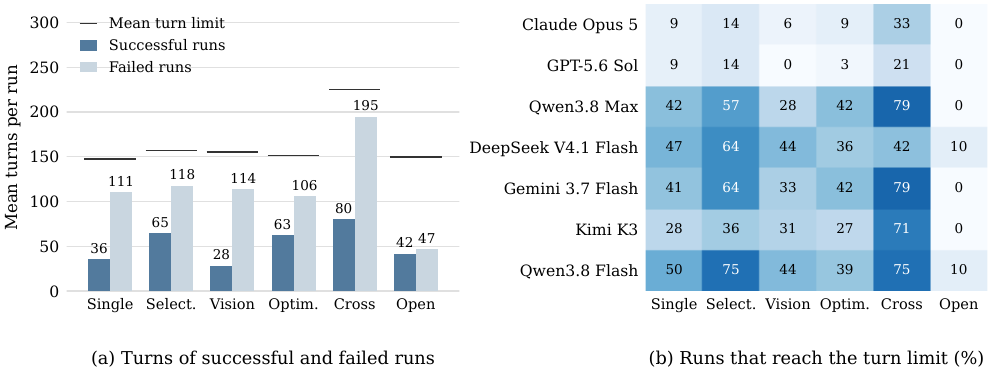}
\caption{\textbf{Turns by task type.} Failed runs use more turns than successful runs in every task type except Open-Environment Engineering.}
\label{fig:turns-by-type}
\end{figure}

The mean turns mainly reflect failed runs (Figure~\ref{fig:turns-by-type}). In every task type except Open-Environment Engineering, failed runs use 1.7 to 4.0 times as many turns as successful runs. Runs that reach the turn limit are most common in Cross-Software Coordination tasks (57\% across all models), and their share ranges from 21\% for GPT-5.6 Sol to 79\% for Gemini 3.7 Flash and Qwen3.8 Max. Failed runs thus consume most of the execution budget. Agents rarely declare failure when they make no progress, so an agent that recognizes a dead end early would save much of this cost.

\begin{table}[!ht]
\centering
\appendixtablestyle
\setlength{\tabcolsep}{4pt}
\caption{\textbf{Execution metrics by task type.} Values are means per task, with turns truncated at the turn limit of the task.}
\label{tab:execution-by-type}
\begin{tabular}{@{}l*{6}{c}@{}}
\toprule
\textbf{Model} & Single & Select. & Vision & Optim. & Cross & Open \\
\midrule
\multicolumn{7}{c}{\textit{Turns}} \\
\midrule
\modellogo{claude}Claude Opus 5 & 57.8 & 89.6 & 48.1 & 72.4 & 176.2 & 28.5 \\
\modellogo{openai}GPT-5.6 Sol & 44.9 & 65.1 & 34.1 & 51.1 & 157.9 & 20.3 \\
\modellogo{qwen}Qwen3.8 Max & 97.5 & 113.2 & 87.8 & 114.4 & 214.2 & 50.7 \\
\modellogo{gemini}Gemini 3.7 Flash & 99.1 & 131.9 & 96.9 & 111.3 & 204.3 & 54.6 \\
\modellogo{deepseek}DeepSeek V4.1 Flash & 108.6 & 136.0 & 110.6 & 108.2 & 184.7 & 48.3 \\
\modellogo{kimi}Kimi K3 & 79.9 & 103.8 & 76.6 & 98.3 & 197.5 & 55.0 \\
\modellogo{qwen}Qwen3.8 Flash & 113.1 & 146.2 & 106.3 & 107.0 & 200.5 & 53.0 \\
\cmidrule(l){2-7}
All models & 85.8 & 112.3 & 80.1 & 94.7 & 190.8 & 44.3 \\
Mean turn limit & 147 & 157 & 156 & 152 & 225 & 150 \\
\midrule
\multicolumn{7}{c}{\textit{Wall-clock time (min)}} \\
\midrule
\modellogo{claude}Claude Opus 5 & 76.8 & 105.3 & 76.8 & 145.7 & 181.1 & 70.2 \\
\modellogo{openai}GPT-5.6 Sol & 48.9 & 77.6 & 63.1 & 72.2 & 139.9 & 17.8 \\
\modellogo{qwen}Qwen3.8 Max & 110.5 & 113.2 & 104.1 & 133.7 & 194.0 & 114.6 \\
\modellogo{gemini}Gemini 3.7 Flash & 41.8 & 50.6 & 40.3 & 53.8 & 84.1 & 14.3 \\
\modellogo{deepseek}DeepSeek V4.1 Flash & 139.5 & 154.5 & 132.9 & 153.2 & 219.8 & 48.6 \\
\modellogo{kimi}Kimi K3 & 52.3 & 57.8 & 110.0 & 71.4 & 111.3 & 55.7 \\
\modellogo{qwen}Qwen3.8 Flash & 53.3 & 68.2 & 47.7 & 58.0 & 97.9 & 49.3 \\
\cmidrule(l){2-7}
All models & 74.7 & 89.6 & 82.1 & 98.3 & 146.9 & 52.9 \\
\midrule
\multicolumn{7}{c}{\textit{API cost (USD)}} \\
\midrule
\modellogo{claude}Claude Opus 5 & 14.10 & 21.15 & 15.14 & 24.94 & 52.13 & 6.88 \\
\modellogo{openai}GPT-5.6 Sol & 7.46 & 10.79 & 5.63 & 9.41 & 32.72 & 2.11 \\
\modellogo{qwen}Qwen3.8 Max & 5.77 & 6.38 & 5.91 & 7.32 & 12.85 & 1.89 \\
\modellogo{gemini}Gemini 3.7 Flash & 1.79 & 2.41 & 1.98 & 2.18 & 4.35 & 0.81 \\
\modellogo{deepseek}DeepSeek V4.1 Flash & 0.76 & 0.88 & 0.75 & 0.79 & 1.38 & 0.22 \\
\modellogo{kimi}Kimi K3 & 4.77 & 6.42 & 5.65 & 6.90 & 13.30 & 2.33 \\
\modellogo{qwen}Qwen3.8 Flash & 0.41 & 0.55 & 0.39 & 0.41 & 0.79 & 0.15 \\
\cmidrule(l){2-7}
All models & 5.01 & 6.94 & 5.06 & 7.42 & 16.79 & 2.06 \\
\bottomrule
\end{tabular}
\end{table}

\subsection{Tasks Missed by All Models}
\label{app:all-zero}

\begin{figure}[!ht]
    \centering
    \includegraphics[width=0.96\linewidth]{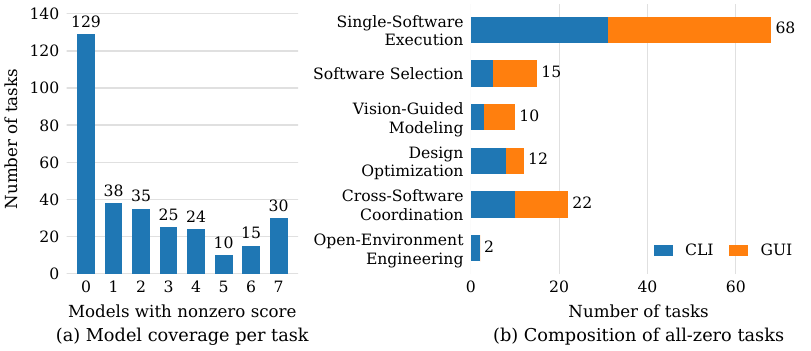}
    \caption{\textbf{Tasks missed by all models.} Cross-Software Coordination has the largest share of such tasks (22 of 24).}
    \label{fig:all-zero-tasks}
\end{figure}

All seven models receive a zero score on 129 of the 306 main-evaluation tasks (Figure~\ref{fig:all-zero-tasks}). Cross-Software Coordination tasks are the hardest by this measure, with 22 of 24 tasks unsolved by every model, followed by Software Selection with 15 of 28. Single-Software Execution still contributes the most all-zero tasks (68), so the gap is not confined to a few difficult cross-software workflows. Only 30 tasks are solved by all seven models, which leaves substantial room to separate future agents.

\subsection{Design Optimization Results}
\label{app:optimization}

Of the 231 main-evaluation runs on the 33 Design Optimization tasks, 169 receive a zero score, 53 a partial score, and 9 a full score (Table~\ref{tab:quantitative-design}). KiCad layout and ZBrush surface tasks account for 38 of the partial scores. Most Design Optimization failures thus come from infeasible designs. Agents seldom reach the quality stage, where the score ranks the designs that pass the feasibility checks (Appendix~\ref{app:scoring}).

\begin{table}[!ht]
\centering
\appendixtablestyle
\caption{\textbf{Design Optimization outcomes.} Most runs receive a zero score because their designs fail the feasibility checks.}
\label{tab:quantitative-design}
\begin{tabular}{@{}lrrrr@{}}
\toprule
 & \multicolumn{3}{c}{\textbf{Number of runs}} & \\
\cmidrule(lr){2-4}
\textbf{Model} & \textbf{Zero} & \textbf{Partial} & \textbf{Full} & \textbf{EngiScore} \\
\midrule
Claude Opus 5 & 21 & 8 & 4 & 32.8 \\
GPT-5.6 Sol & 20 & 8 & 5 & 36.4 \\
Qwen3.8 Max & 24 & 9 & 0 & 22.3 \\
Gemini 3.7 Flash & 24 & 9 & 0 & 20.6 \\
DeepSeek V4.1 Flash & 27 & 6 & 0 & 16.2 \\
Kimi K3 & 24 & 9 & 0 & 21.9 \\
Qwen3.8 Flash & 29 & 4 & 0 & 9.9 \\
\midrule
All runs & 169 & 53 & 9 & 22.9 \\
\bottomrule
\end{tabular}
\end{table}

\paragraph{Quality differences among valid outputs.}
On ZBrush task~04, Gemini 3.7 Flash and GPT-5.6 Sol both produce valid surfaces with full evaluation-grid coverage. GPT-5.6 Sol achieves a lower height-field RMSE (0.077 versus 0.151 OBJ units) and a higher score (0.892 versus 0.804; Table~\ref{tab:quantitative-metrics}). Since the initial design has a score of 0.880, GPT-5.6 Sol improves it slightly, whereas Gemini 3.7 Flash returns a valid but lower-quality surface.

\paragraph{Interpreting a near-full score.}
On KiCad task~01, Gemini 3.7 Flash receives a score of 0.982, compared with 0.647 for the baseline layout. The weighted layout cost falls from 64.21 to 3.21 and the route length from 53.16 to 40.17\,mm. All four movable components reach their target positions, with no vias and a minimum spacing of 4.62\,mm against a 2.0\,mm requirement. The remaining cost comes from the route-length term, so the layout meets every requirement and loses score only on route length.

\paragraph{Feasibility gates and zero scores.}
On the same ZBrush task~04, Qwen3.8 Flash obtains an RMSE of 0.181 OBJ units but receives a zero score, because its mesh contains degenerate faces, overlapping projected triangles, and multiple surface heights at the sampling points. GPT-5.6 Sol also receives a zero score on ZBrush task~05, where its grid coverage of 94.75\% falls below the required 98\%. Approximation accuracy cannot compensate for invalid topology or insufficient coverage.

\begin{table*}[!ht]
\centering
\appendixtablestyle
\caption{\textbf{Representative Design Optimization outcomes.} Raw quality measurements behind selected scores.}
\label{tab:quantitative-metrics}
\begin{tabular}{@{}lrrrrrr@{}}
\toprule
\multicolumn{7}{@{}l}{\textbf{KiCad $\mid$ task 01: Layout optimization}} \\
Model & Cost & Route (mm) & Spacing (mm) & Vias & Baseline & Score \\
\cmidrule(lr){1-5}\cmidrule(l){6-7}
Gemini 3.7 Flash & 3.21 & 40.17 & 4.62 & 0 & 0.647 & 0.982 \\
\cmidrule(lr){1-7}
\multicolumn{7}{@{}l}{\textbf{ZBrush $\mid$ task 04: Thumb-rest surface}} \\
Model & RMSE & Coverage (\%) & Faces & Valid & Baseline & Score \\
\cmidrule(lr){1-5}\cmidrule(l){6-7}
Gemini 3.7 Flash & 0.151 & 100.00 & 256 & Pass & 0.880 & 0.804 \\
GPT-5.6 Sol & 0.077 & 100.00 & 256 & Pass & 0.880 & 0.892 \\
Qwen3.8 Flash & 0.181 & 100.00 & 24,576 & Fail & 0.880 & 0.000 \\
\cmidrule(lr){1-7}
\multicolumn{7}{@{}l}{\textbf{ZBrush $\mid$ task 05: Procedural ridge surface}} \\
Model & RMSE & Coverage (\%) & Faces & Valid & Baseline & Score \\
\cmidrule(lr){1-5}\cmidrule(l){6-7}
GPT-5.6 Sol & 0.493 & 94.75 & 256 & Fail & 0.894 & 0.000 \\
\bottomrule
\end{tabular}
\end{table*}

\subsection{Ablation Details}
\label{app:ablations}

\paragraph{Setup.}
All ablations use GPT-5.6 Sol, Gemini 3.7 Flash, and Kimi K3, and change one factor while keeping all other settings identical to those in the main evaluation (Table~\ref{tab:ablation-setup}). The initial-image ablation uses 60 Vision-Guided Modeling tasks per interface. Its CLI arm keeps the image-reading tool enabled, so only the location of the reference drawing changes. The runtime image-access ablation is CLI only and toggles the image-reading tool on 60 tasks whose reference drawing is placed in the task message and on 60 tasks without an initial drawing, where the tool can only be used to inspect renderings of the agent's own intermediate artifacts. The accessibility and resolution ablations use the same 100 GUI tasks. The resolutions of $0.2\times$, $0.6\times$, and $1.0\times$ correspond to $384\times216$, $1152\times648$, and $1920\times1080$ screenshots. The history ablation uses 100 tasks spanning both interfaces. Conditions identical to another configuration, such as the $1.0\times$ resolution and the plain-screenshot condition, share runs and are counted once.

\begin{table}[!ht]
\centering
\appendixtablestyle
\caption{\textbf{Ablation setup.}}
\label{tab:ablation-setup}
\begin{tabular}{@{}llll@{}}
\toprule
\textbf{Ablation} & \textbf{Interface} & \textbf{Conditions} & \textbf{Tasks} \\
\midrule
Initial-image presentation & GUI, CLI & Env., Msg. & 60 per interface \\
Runtime image access & CLI & Tool off, on; with or without initial drawing & 60 per drawing setting \\
Accessibility information & GUI & Screenshot, Screenshot + A11y & 100 \\
Screenshot resolution & GUI & $0.2\times$, $0.6\times$, $1.0\times$ & 100 \\
History length & GUI, CLI & 5, 10, 15 turns & 100 \\
\bottomrule
\end{tabular}
\end{table}

\paragraph{Initial-image presentation.}
Presenting the reference drawing in the message raises the GUI EngiScore of all three models, while the CLI score of Kimi K3 falls from 40.0 to 36.7 (Table~\ref{tab:initial-image}). For GPT-5.6 Sol, 15 GUI tasks turn from failure to success and five from success to failure. FreeCAD task~01, which becomes successful, needs dimensions from the drawing and four hole locations from a supplement, so presenting the drawing up front removes one retrieval step. Blender task~02, which becomes unsuccessful, requires a wall of 10 courses of 10 bricks modeled as 100 disconnected islands in one mesh. Recognizing the wall does not ensure this mesh structure. Easier access to a reference thus helps agents gather information but does not guarantee a structurally correct artifact.

\paragraph{Runtime image access in CLI.}
Enabling the image-reading tool improves Gemini 3.7 Flash on both task sets and lowers Kimi K3 on both (Table~\ref{tab:cli-visual}). For GPT-5.6 Sol, it raises EngiScore from 66.7 to 68.3 with an initial drawing but lowers it from 48.1 to 41.5 without one. With the drawing in the message, Kimi K3 gains two tasks and loses five. For example, FreeCAD task~39 becomes successful while task~42 fails. Both tasks require a single valid solid with bores, slots, or hole patterns at specified depths and angles. A rendered view lets the agent compare these features with the reference drawing, but it cannot confirm that their depths and angles are exact. Rendered views therefore help agents check the shape of a feature but not its exact dimensions.

\paragraph{Accessibility information.}
Adding an accessibility tree improves Kimi K3, lowers Gemini 3.7 Flash, and slightly lowers GPT-5.6 Sol, from 29.8 to 29.1 (Figure~\ref{fig:observation-history}a). For GPT-5.6 Sol, eight tasks gain a nonzero score and eight lose one, but the losses (8.0 points) outweigh the gains (7.2 points). Within FreeCAD, task~14, which fuses a cylindrical crossbar between two posts, becomes successful, while task~20, which positions and fuses an upright support and then cuts a through-hole along a given axis, fails. Interface text helps locate controls, but these operations still depend on selecting the right body, placement, and feature direction.

\paragraph{Screenshot resolution.}
Native resolution gives the highest EngiScore for all three models (Figure~\ref{fig:observation-history}b). For GPT-5.6 Sol, moving from $0.6\times$ to $1.0\times$ yields 10 new full-score tasks and 10 regressions costing 9.1 points, a small net gain from 29.0 to 29.8. Kimi K3 gains 11 tasks and loses none. The newly solved KiCad task~06 requires repairing the exposed pad, paste apertures, and courtyard of a QFN-48 footprint, where finer detail helps distinguish small features. The newly failed Blender task~02 is the brick-wall task above, whose difficulty lies in the mesh structure. Finer detail therefore helps on tasks with small visual features but not on tasks limited by model structure.

\paragraph{History length.}
Extending the history from 10 to 15 turns lowers Kimi K3 from 26.2 to 20.1 and leaves GPT-5.6 Sol nearly unchanged, from 36.6 to 36.7 (Figure~\ref{fig:observation-history}c). The decline for Kimi K3 comes mainly from tasks whose score drops to zero: five tasks gain a nonzero score (+4.4 points) and 11 lose it ($-$10.5 points). For GPT-5.6 Sol, seven gains (+6.9) and seven full-score losses ($-$7.0) nearly cancel, while ZBrush task~03 improves from 0.614 to 0.844, a quality change that a success count would miss. This score still falls below the 0.909 of the initial design, so the longer history improves the output without surpassing the starting design.

% !TEX root = ../main.tex
\FloatBarrier
\section{Case Studies and Failure Analysis}
\label{app:cases}

\subsection{Successful Workflows}
\label{app:success}

\begin{figure}[!ht]
    \centering
    \includegraphics[width=\linewidth]{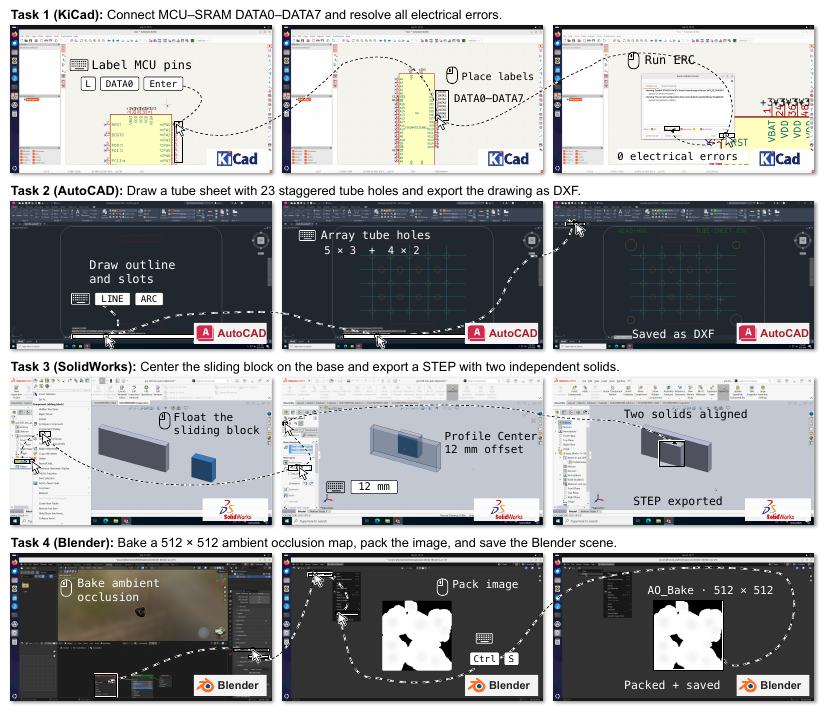}
    \caption{\textbf{Successful Single-Software Execution workflows.} Runs in KiCad, AutoCAD, SolidWorks, and Blender on the GUI interface.}
    \label{fig:single-software-workflows}
\end{figure}

Figure~\ref{fig:single-software-workflows} shows four successful Single-Software Execution runs on the GUI interface, all of which pass their verifiers with a score of 1.0. In all four runs, the agent completes the requested change, keeps the existing design content, and saves the result into the delivered file, which the verifier reopens.

\paragraph{KiCad.} The agent connects the MCU and SRAM through eight net labels, \texttt{DATA0}--\texttt{DATA7}, that establish the required one-to-one pin mapping, and adds the two required power flags while keeping the existing components and power and control connections. The saved schematic contains both the new data connections and the original circuitry, with no electrical-rule errors.

\paragraph{AutoCAD.} After drawing the outline and slots, the agent combines $5 \times 3$ and $4 \times 2$ rectangular arrays with offset origins to produce 23 staggered tube holes. The exported DXF retains the required coordinates, radii, entity types, and layer assignments.

\paragraph{SolidWorks.} The agent makes the sliding block movable, applies a Profile Center mate, and corrects the offset direction so that the block is centered on the base and its bottom face rests on the top face of the base. The exported STEP satisfies these placement constraints while keeping the original dimensions and the two separate solids.

\paragraph{Blender.} The agent bakes ambient occlusion into a $512 \times 512$ image and packs it into the scene before saving, so the baked image is stored in the delivered file.

\subsection{Failure Statistics}
\label{app:failure-stats}

We group zero-score runs by terminal outcome into five categories: declared completion, reaching the turn limit, declared failure, runtime-limit violation, and unexpected desktop closure or restart. Runs shared between the main evaluation and an ablation with the same configuration are counted once, and 10 failures whose records name an evaluator-side cause, a missing output file or required entry, are excluded (7 of them in the main evaluation). Table~\ref{tab:failure-taxonomy} covers the 3,366 analyzed failures across the main evaluation and ablations. Table~\ref{tab:failure-main} and Figure~\ref{fig:failure-reasons} break down the 1,506 main-evaluation failures by model, and Figure~\ref{fig:interface-failure-breakdown} by interface. Reaching the turn limit accounts for 35.9\% of CLI failures (235 of 655) and 55.3\% of GUI failures (471 of 851), whereas declared completion accounts for 60.6\% of CLI failures and 37.0\% of GUI failures.

\begin{table}[!ht]
\centering
\appendixtablestyle
\caption{\textbf{Failure categories across all experiments.} Counts and shares of the 3,366 analyzed failed runs.}
\label{tab:failure-taxonomy}
\begin{tabular}{@{}lrr@{}}
\toprule
\textbf{Recorded outcome} & \textbf{Runs} & \textbf{Share (\%)} \\
\midrule
Declared completion & 1,658 & 49.26 \\
Turn limit reached & 1,501 & 44.59 \\
Declared failure & 167 & 4.96 \\
Runtime limit exceeded & 37 & 1.10 \\
Unexpected desktop closure or restart & 3 & 0.09 \\
\bottomrule
\end{tabular}
\end{table}

\begin{table}[!ht]
\centering
\appendixtablestyle
\caption{\textbf{Failure counts by model.} Analyzed failed runs in the 306-task main evaluation.}
\label{tab:failure-main}
\begin{tabular}{@{}lrrrrrr@{}}
\toprule
\textbf{Model} & \textbf{Analyzed} & \textbf{Completion} & \textbf{Turn limit} & \textbf{Failure} & \textbf{Runtime limit} & \textbf{Desktop} \\
\midrule
GPT-5.6 Sol & 186 & 162 & 24 & 0 & 0 & 0 \\
Claude Opus 5 & 167 & 124 & 32 & 2 & 9 & 0 \\
Gemini 3.7 Flash & 224 & 90 & 133 & 0 & 0 & 1 \\
Kimi K3 & 230 & 126 & 94 & 10 & 0 & 0 \\
Qwen3.8 Max & 224 & 87 & 131 & 4 & 0 & 2 \\
Qwen3.8 Flash & 254 & 68 & 155 & 31 & 0 & 0 \\
DeepSeek V4.1 Flash & 221 & 55 & 137 & 1 & 28 & 0 \\
\midrule
Total & 1,506 & 712 & 706 & 48 & 37 & 3 \\
\bottomrule
\end{tabular}
\end{table}

\subsection{Verified Violations after Declared Completion}
\label{app:violations}

Table~\ref{tab:artifact-violations} lists four runs that declare completion but fail verification. In each case, the verifier finds a geometric or cross-stage violation in the delivered artifacts, even when the report written by the agent claims that all requirements are met.

\begin{table}[!ht]
\centering
\appendixtablestyle
\caption{\textbf{Verified violations after declared completion.} Each run ends with a completion declaration but receives a zero score because the submitted artifact violates a task requirement.}
\label{tab:artifact-violations}
\begingroup
\begin{tabularx}{\linewidth}{@{}>{\raggedright\arraybackslash}p{0.27\linewidth}>{\raggedright\arraybackslash}p{0.19\linewidth}>{\raggedright\arraybackslash}X@{}}
\toprule
\textbf{Task} & \textbf{Model} & \textbf{Verified violation} \\
\midrule
OpenSCAD microfluidic lid (GUI) & Gemini 3.7 Flash & Channel cut from the wrong face; sampled material and cavity states are inverted. \\
OpenSCAD V-groove pulley (CLI) & Gemini 3.7 Flash & Groove root and section fail at the required stations in the inherited datum frame. \\
KiCad--OpenSCAD--FreeCAD--Blender (CLI) & Gemini 3.7 Flash & The mechanical map assigns a mounting-hole role inconsistent with the source board. \\
Thermal-camera mount (Open-Environment Engineering, CLI) & GPT-5.6 Sol & The MH1 bore contains $3.70$\,mm$^3$ of residual material, exceeding the $0.1$\,mm$^3$ allowance. \\
\bottomrule
\end{tabularx}
\endgroup
\end{table}

\subsection{Cross-Software Coordination Failure Cases}
\label{app:cross-failures}

\begin{figure}[!ht]
    \centering
    \includegraphics[width=\linewidth]{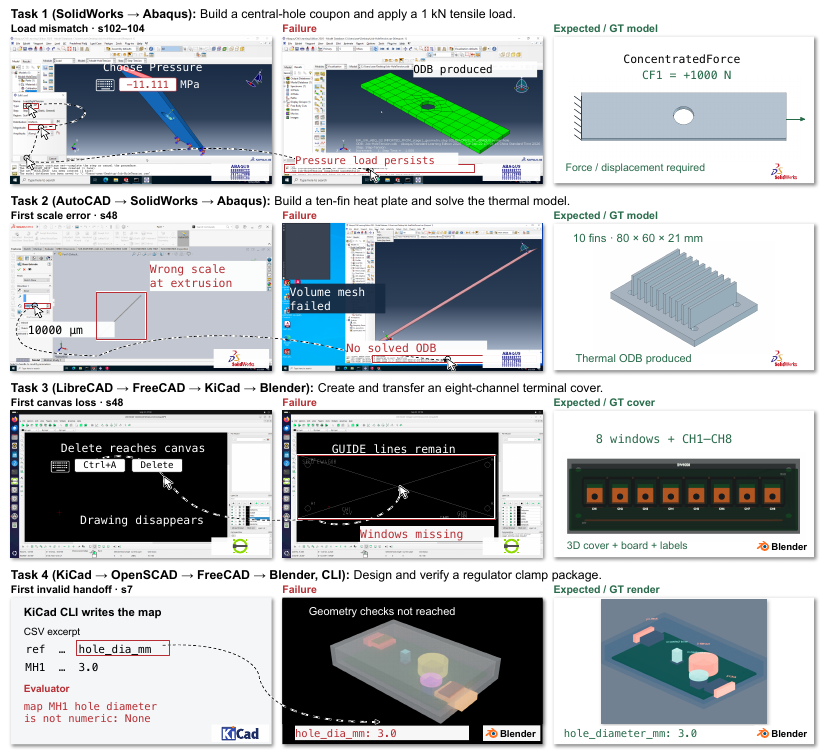}
    \caption{\textbf{Cross-Software Coordination failure cases.} Two runs fail at the handoffs between applications, and two fail inside one application.}
    \label{fig:multi-software-failures}
\end{figure}

Figure~\ref{fig:multi-software-failures} compares four failed Cross-Software Coordination runs with renderings of the reference solutions. Two of the failures occur at the handoffs between applications, and two occur inside one application. Two of the runs declare completion, one reaches the 300-turn limit, and one declares failure.

\paragraph{Load representation mismatch.}
In the SolidWorks--Abaqus case, the analysis completes and produces an output database, but the model applies the load as a $-11.111$\,MPa pressure. Although this pressure corresponds to about 1\,kN of tension on the specified $30 \times 3$\,mm end face, the task requires the load to be defined as a native force or displacement parameter, so the submission violates the load specification and fails the load check.

\paragraph{Import scale errors.}
In the AutoCAD--SolidWorks--Abaqus case, an import-scale error becomes visible during extrusion. The agent corrects the units, but the error returns after a crash and a second DXF import. The transferred model is elongated and lacks the proportions of the $80 \times 60 \times 21$\,mm 10-fin reference, and repeated meshing attempts continue until the 300-turn limit without a solved thermal model.

\paragraph{Incorrect action targeting.}
The LibreCAD--FreeCAD--KiCad--Blender case stalls during drawing preparation. An intended text edit instead selects and deletes every object on the drawing canvas. After the seed drawing is reopened, the saved DXF still contains the diagonal guide lines and lacks the eight required windows. Later attempts to launch FreeCAD do not advance the workflow, and the run ends with a failure declaration.

\paragraph{Column-name mismatch.}
The KiCad--OpenSCAD--FreeCAD--Blender CLI case produces all 18 required files, including the final Blender render, and declares completion. However, its mechanical map does not use the column name that the task specifies for the mounting-hole diameter, so the verifier cannot read the diameter of hole MH1 and stops before the downstream geometry checks.

\subsection{Cross-Software Coordination versus Open-Environment Engineering}
\label{app:paired}

The ten Open-Environment Engineering tasks in the main evaluation are derived from the ten KiCad--OpenSCAD--FreeCAD--Blender Cross-Software Coordination tasks, and each pair shares the same engineering target (Figure~\ref{fig:shared-task-cases} shows tasks 02 and 09). The Cross-Software Coordination versions of tasks 02, 04, 05, and 09 are also in the main evaluation, which allows a paired comparison on these four tasks.

\begin{figure}[!ht]
\centering
\begin{tcolorbox}[
  enhanced,
  colback=gray!8,
  colframe=gray!60!black,
  coltitle=white,
  fontupper=\fontfamily{ptm}\selectfont,
  fonttitle=\bfseries,
  title=Shared task cases,
  boxrule=0.55pt,
  arc=2pt,
  left=6pt,
  right=6pt,
  top=5pt,
  bottom=5pt,
  before skip=5pt,
  after skip=8pt
]
\small
\textbf{Task 02 \quad Buck-regulator heat-spreader clamp package}\par\smallskip
\begin{tabularx}{\linewidth}{@{}>{\bfseries}p{0.20\linewidth}X@{}}
Task type & Cross-Software Coordination and Open-Environment Engineering \\
Shared design & PA12 enclosure for a buck-regulator PCB with thermal contact between U1 and the lid. \\
Geometry & $117.2 \times 75.2 \times 18.6$\,mm package; base, side walls, lid, and four bored PCB standoffs; $6 \times 6$\,mm contact column. \\
Hard constraints & Contact column within 0.1\,mm of U1; empty L1 exclusion region; at least 2.4\,mm clearance above C1; side-wall openings for J1 and J2. \\
\end{tabularx}
\par\medskip
\noindent\rule{\linewidth}{0.35pt}\par\medskip
\textbf{Task 09 \quad Thermal-camera calibration-target PCB mount}\par\smallskip
\begin{tabularx}{\linewidth}{@{}>{\bfseries}p{0.20\linewidth}X@{}}
Task type & Cross-Software Coordination and Open-Environment Engineering \\
Shared design & Black-PA12 mount for a calibration-target PCB with two exposed heaters. \\
Geometry & $131 \times 105 \times 25.1$\,mm assembly; tray, separate lid, and four bored PCB standoffs; two independent 36\,mm openings above H1 and H2. \\
Hard constraints & Preserve PCB placement and connector clearances; keep J1 and J2 accessible; each heater opening at least 99\% open with an edge gap of at least 10\,mm. \\
\end{tabularx}
\end{tcolorbox}
\caption{\textbf{Paired Cross-Software Coordination and Open-Environment Engineering tasks.} Tasks 02 and 09 share engineering targets but differ in workflow and delivery requirements.}
\label{fig:shared-task-cases}
\end{figure}

\begin{table}[!ht]
\centering
\appendixtablestyle
\setlength{\tabcolsep}{2.5pt}
\caption{\textbf{Paired task outcomes.} No model solves a Cross-Software Coordination version, while 13 of 28 Open-Environment Engineering runs succeed.}
\label{tab:challenge-results}
\begin{tabular}{@{}lrrrrrrrr@{}}
\toprule
 & \multicolumn{2}{c}{\textbf{Task 02}} & \multicolumn{2}{c}{\textbf{Task 04}} & \multicolumn{2}{c}{\textbf{Task 05}} & \multicolumn{2}{c}{\textbf{Task 09}} \\
\cmidrule(lr){2-3}\cmidrule(lr){4-5}\cmidrule(lr){6-7}\cmidrule(l){8-9}
\textbf{Model} & Cross & Open & Cross & Open & Cross & Open & Cross & Open \\
\midrule
GPT-5.6 Sol & 0 & 0 & 0 & 1 & 0 & 1 & 0 & 0 \\
Claude Opus 5 & 0 & 0 & 0 & 1 & 0 & 1 & 0 & 1 \\
Gemini 3.7 Flash & 0 & 0 & 0 & 1 & 0 & 1 & 0 & 1 \\
Kimi K3 & 0 & 0 & 0 & 0 & 0 & 0 & 0 & 0 \\
Qwen3.8 Max & 0 & 0 & 0 & 0 & 0 & 1 & 0 & 1 \\
Qwen3.8 Flash & 0 & 0 & 0 & 0 & 0 & 1 & 0 & 0 \\
DeepSeek V4.1 Flash & 0 & 0 & 0 & 0 & 0 & 1 & 0 & 1 \\
\bottomrule
\end{tabular}
\end{table}

\paragraph{Same design, different deliverables.}
The Cross-Software Coordination version prescribes a sequence: KiCad extracts the board geometry and parameters, OpenSCAD builds the enclosure, FreeCAD assembles the parts and checks clearances, and Blender renders a visual review. It requires 18 artifacts, including intermediate handoffs, reports, a rendered review, and a toolchain invocation log. The Open-Environment Engineering version allows any software, scripts, or libraries in any order and requires only one STEP assembly of the enclosure, PCB, components, and task-specific features, which is checked against the geometric requirements.

\paragraph{Outcomes.}
All seven models fail the Cross-Software Coordination versions of all four tasks (Table~\ref{tab:challenge-results}). On the Open-Environment Engineering versions, six models succeed on task~05, four on task~09, and three on task~04, while task~02 remains unsolved by every model. In the Cross-Software Coordination version of task~09, Gemini 3.7 Flash submits a toolchain invocation log of 478 bytes, below the 500-byte minimum, after 44 of its 150 turns. Qwen3.8 Max reaches its 150-turn limit with the same file missing. In the Open-Environment Engineering version, both models build the assembly through FreeCAD scripting and pass the geometry checks, Gemini 3.7 Flash in 41 turns and Qwen3.8 Max in 30. Both Cross-Software Coordination runs thus fail on delivery requirements before their geometry is checked. Geometric errors still occur when the workflow is free: GPT-5.6 Sol leaves 3.70\,mm$^3$ of material in the MH1 bore of task~09, above the 0.1\,mm$^3$ tolerance. The paired results separate two demands of Cross-Software Coordination tasks: building a valid design and completing the required handoffs and documentation.

\end{document}